\pdfoutput=1

\documentclass[11pt]{article}
\usepackage{arydshln}
\usepackage{emnlp2022}
\usepackage{booktabs}
\usepackage{multirow}
\usepackage{times}
\usepackage{latexsym}
\usepackage{amsmath}
\usepackage{amssymb}
\usepackage{enumitem}
\usepackage{svg}
\usepackage{adjustbox}
\usepackage[T1]{fontenc}
\usepackage{float}
\usepackage[utf8]{inputenc}
\usepackage{microtype}
\usepackage{tipa}
\usepackage{graphicx}

\title{Bootstrapping meaning through listening: Unsupervised learning of spoken sentence embeddings}

\author{Jian Zhu\textsuperscript{\textipa{B},\textipa{\oe}}, Zuoyu Tian\textsuperscript{\textipa{X}}, Yadong Liu\textsuperscript{\textipa{\oe}}, Cong Zhang\textsuperscript{\textipa{Q}}, Chia-wen Lo\textsuperscript{\textipa{M}} \\
    \textsuperscript{\textbf{\textipa{B}}}University of Michigan, Ann Arbor \hspace{0.1in} 
    \textsuperscript{\textbf{\textipa{\oe}}}University of British Columbia \\
    \textsuperscript{\textbf{\textipa{X}}}Indiana University Bloomington \hspace{0.3in} 
    \textsuperscript{\textbf{\textipa{Q}}}Newcastle University \\
    \textsuperscript{\textbf{\textipa{M}}}Max Planck Institute for Human Cognitive and Brain Sciences \\
    \textsuperscript{\textbf{\textipa{B}}}\texttt{lingjzhu@umich.edu}, \textsuperscript{\textbf{\textipa{X}}}\texttt{zuoytian@iu.edu}
       }

\begin{document}
\maketitle
\vspace{0.1in}
\begin{abstract}
Inducing semantic representations directly from speech signals is a highly challenging task but has many useful applications in speech mining and spoken language understanding. This study tackles the unsupervised learning of semantic representations for spoken utterances. Through converting speech signals into hidden units generated from acoustic unit discovery, we propose WavEmbed, a multimodal sequential autoencoder that predicts hidden units from a dense representation of speech. Secondly, we also propose S-HuBERT to induce meaning through knowledge distillation, in which a sentence embedding model is first trained on hidden units and passes its knowledge to a speech encoder through contrastive learning. The best performing model achieves a moderate correlation (0.5$\sim$0.6) with human judgments, without relying on any labels or transcriptions. Furthermore, these models can also be easily extended to leverage textual transcriptions of speech to learn much better speech embeddings that are strongly correlated with human annotations. Our proposed methods are applicable to the development of purely data-driven systems for speech mining, indexing and search.
\end{abstract}

\section{Introduction}
In Spoken Language Understanding (SLU), a goal is to understand the semantic content of spoken utterances. Traditionally, research in speech processing focus on tasks that process the low-level sensory information in speech, such as automatic speech processing (ASR), under the assumption that language understanding can be handled by NLP modules after speech is transcribed \citep{wang2005spoken,de2008spoken,serdyuk2018towards}. Yet speech-based semantic representations allow us to bypass texts in some scenarios, not only simplifying the pipeline and but also beneficial for certain domains without much transcribed data or some languages without writing systems. 

For speech processing, the spoken term detection tasks such as keyword detection \cite[e.g.,][]{mamou2007vocabulary,miller07_interspeech,5752829,wang2018segmental} and query-by-example search \cite[e.g.,][]{hazen2009query,parada2009query,chen2015query} focus on the exact matching of audio terms in speech databases. Yet the speech-to-speech search enabled by spoken sentence embeddings can further expand our capacity to search speech in meaning rather than only in form. This capacity marks a significant advancement in the machine's capacity to perform speech mining, voice search and indexing and spoken information retrieval \cite{duquenne2021multimodal}. 

\begin{figure*}[htbp]
  \centering
  \adjustbox{trim=4cm 4.5cm 5cm 1cm}{%
    \includegraphics[width=\textwidth]{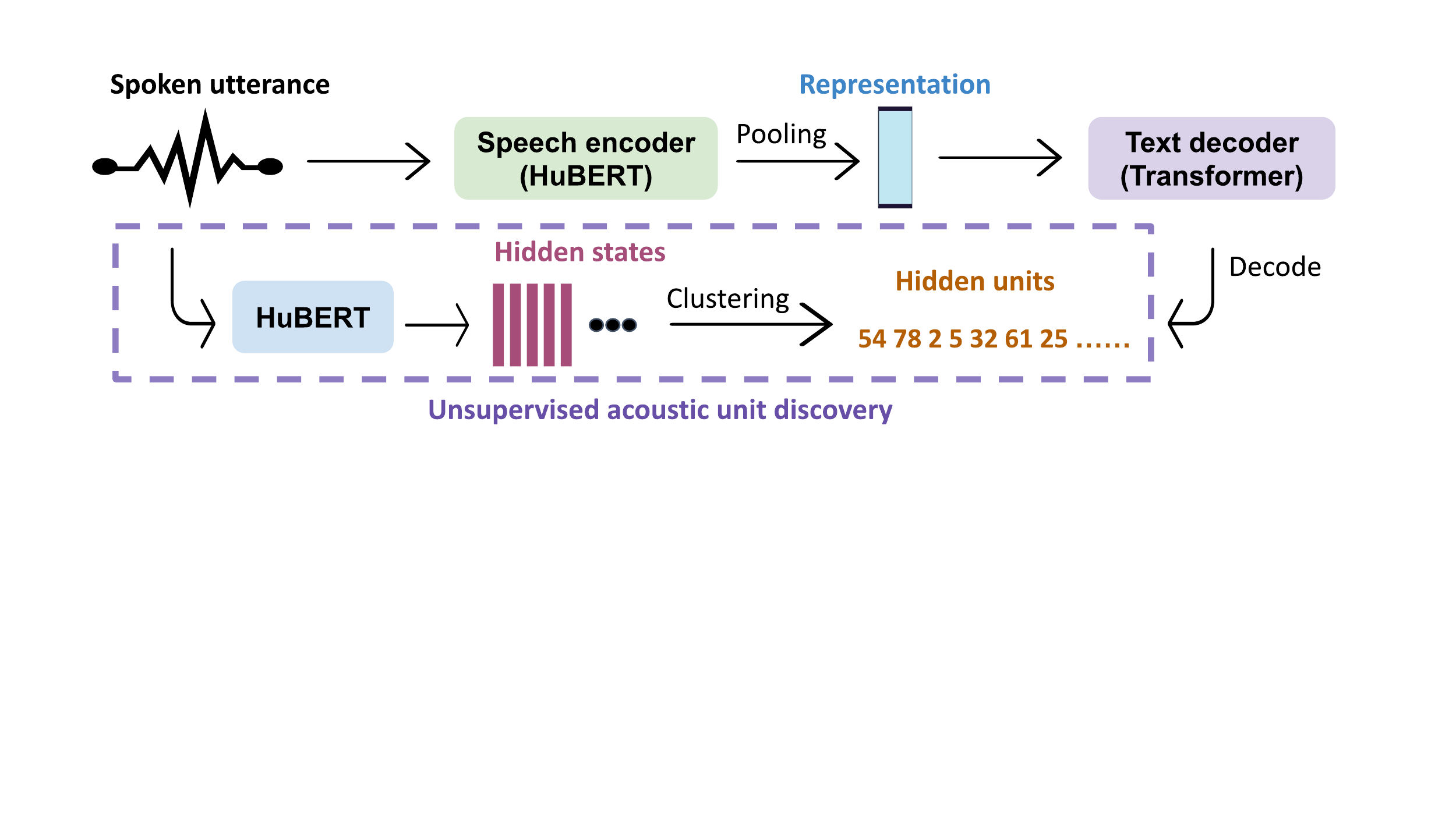}
  }
  \caption{The architecture of WavEmbed. WavEmbed first projects a speech signal into a fixed-dimensional vector representation, and then decodes it back to discrete acoustic units, which are generated through clustering on the hidden states from the sixth layer of the (frozen) pretrained HuBERT model. The learned fixed-dimensional vector encodes semantic information in the latent space.  No texts are required in this training loop. However, if text transcripts are available, the decoder targets can also be textual sequences.}
  \label{fig:wavembed_archi}
\end{figure*}

\begin{figure}[t]
  \centering
  \adjustbox{trim={4.5cm 2cm 5.1cm 2.2cm},clip}{%
    \includegraphics[width=\textwidth]{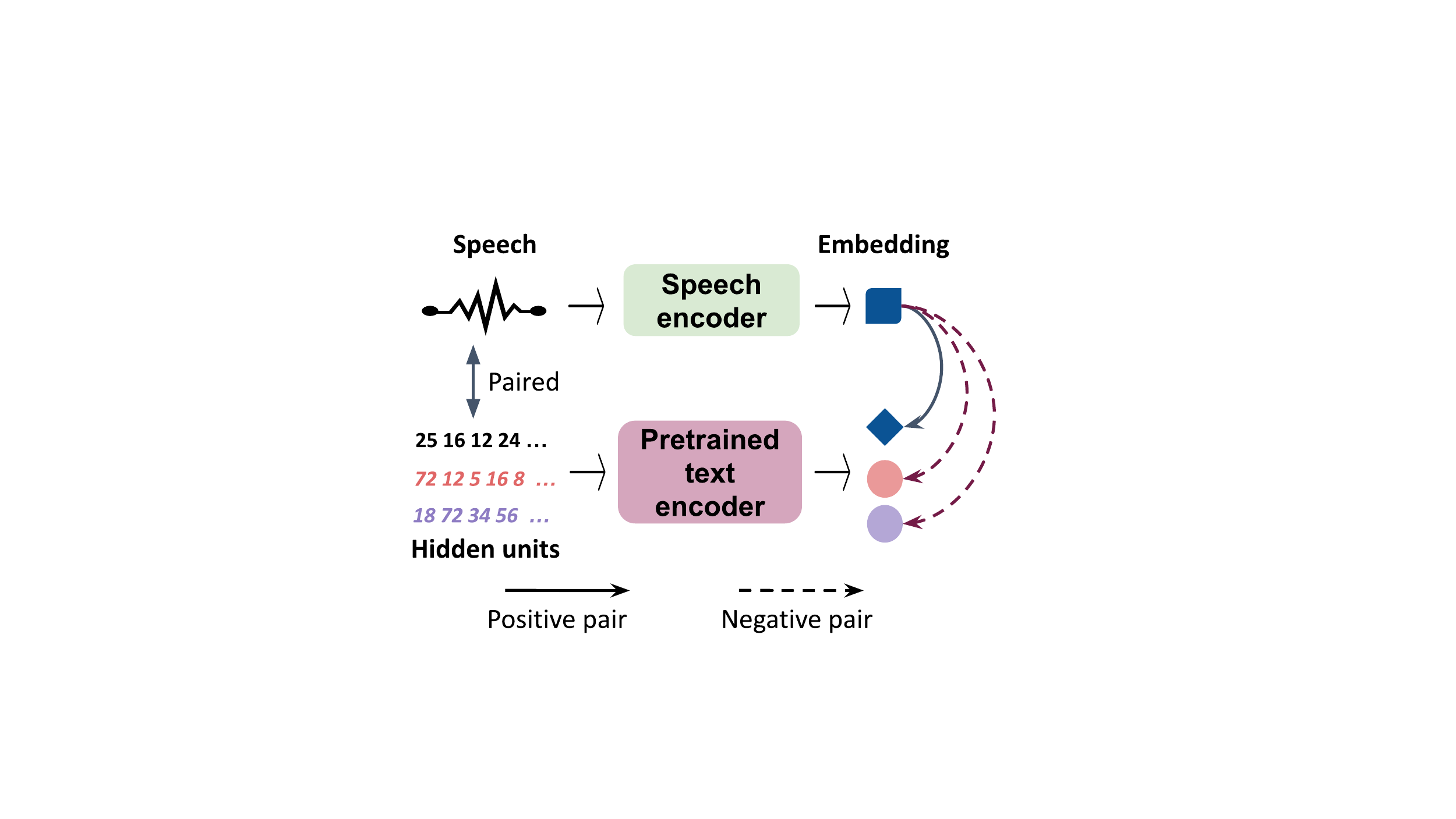}
  }
  \caption{Illustration of S-HuBERT. An unsupervised text embedding model is first trained on either hidden units or textual transcripts. Then it is used as a teacher model to transfer semantic knowledge to a speech encoder through contrastive model distillation.}
  \label{fig:distil}
\end{figure}

While learning textual sentence similarity is a classic task in NLP \citep[e.g.,][]{agirre-etal-2012-semeval,agirre-etal-2015-semeval,agirre-etal-2016-semeval,cer-etal-2017-semeval}, the task is still relatively unexplored in speech research. The main challenge in learning spoken sentence embeddings lies in the lack of labeled data for supervised learning. 
Given the costs associated by creating semantic ratings, it is important to explore unsupervised methods to induce semantic representations directly from speech signals. 

In this study, we present two approaches to tackle the challenge of inducing semantic representations directly from speech signals without any semantic labeling. The first model, Waveform Embedding Transformer (WavEmbed) (Figure~\ref{fig:wavembed_archi}), is a multimodal sequential autoencoder that encodes a speech signal into a bottleneck vector and reconstructs a sequence of  `\textit{hidden units}', which are generated using unsupervised acoustic unit discovery. The second model, Sentence HuBERT (S-HuBERT), learns the semantic representation through aligning with a frozen unsupervised text embedding model, which is trained with the hidden units (Figure~\ref{fig:distil}). We make the following contributions.

\begin{itemize}
    \item We propose simple yet effective unsupervised methods to learn spoken sentence representations. Our best performing unsupervised model achieves moderate Spearman's rank correlations (0.5$\sim$0.6) with human judgements without relying on any labels or text transcriptions.
    
    \item Our proposed methods can be easily extended to speech-text pairs to enhance performance. With text transcriptions, the performance can further be increased to 0.7$\sim$0.8 in terms of Spearman's correlation. We made extensive comparisons and analyses of model performance under different conditions. 
    
    \item We have also created a speech dataset for evaluating spoken sentence similarities, which were rated by multiple human raters and encompassed various speech accents to measure the robustness of models.
\end{itemize}


Our code, data and pretrained checkpoints necessary for replicating the experiments are available at \url{https://github.com/lingjzhu/spoken_sent_embedding}.

\section{Background}
\paragraph{Self-supervised speech modeling}
Most speech technologies including ASR and text-to-speech synthesis (TTS) nowadays heavily rely on the availability of text transcripts. Yet such textual resources can sometimes be hard to collect for many languages, some of which might not have writing systems. Many efforts have since been made to explore effective methods to learn speech representations directly from speech signals, such as the ZeroSpeech Workshop \cite{versteegh15_interspeech,dunbar2017zero,Dunbar2019,Dunbar2020,dunbar21_interspeech}. 

Recently, large-scale self-supervised models including CPC \cite{oord2018representation}, Wav2Vec \cite{Schneider2019}, Wav2Vec2 \cite{baevski2020wav2vec}, HuBERT \cite{hsu2021hubert} and WavLM \cite{chen2021wavlm} have learned effective speech representations that can benefit a wide range of downstream speech tasks \cite{yang21c_interspeech}. In particular, \citet{hsu2021hubert} proposed using clustering algorithm to cluster hidden states of HuBERT into hidden units, which were then used to create training masks. These clusters were shown to encode rich phonemic information \cite{hsu2021hubert,baevski2021unsupervised}. Later it is found that discretizing speech into `hidden units' allows the application of NLP algorithms to process speech via the proxy of these discrete hidden units, without the need of actual textual transcriptions (`textless NLP') \cite{lakhotia-etal-2021-generative,nguyen2022discrete}. This discovery has greatly benefited a variety of tasks, some of which were traditionally not performed with speech, including unsupervised ASR \cite{baevski2021unsupervised}, 
spoken language modeling \cite{ao2022pre,hsu2021hubert,wu2022wav2seq,nguyen2022discrete}, speech resynthesis \cite{polyak2021speech}, spoken language generation \cite{lakhotia-etal-2021-generative,kharitonov-etal-2022-text}, speech-to-speech translation \cite{lee2021textless,popuri2022enhanced,lee-etal-2022-direct,wu2022wav2seq}, and spoken named entity recognition \cite{wu2022wav2seq}. Following these prior studies, our work also falls in the domain of `textless NLP'.

\paragraph{Unsupervised sentence embeddings}
Learning semantic sentence embedding has been extensively studied in NLP community, 
such as Skip-Thought vectors \cite{kiros2015skip}, InferSent \cite{conneau-etal-2017-supervised}, Universal Sentence Encoder \cite{cer2018universal}, and SBERT \cite{reimers-gurevych-2019-sentence}. Recently, unsupervised sentence embeddings have considerably narrowed the performance gap between unsupervised and supervised methods. Contrastive learning has been utilized to learn a vector space in which semantically similar sentences are close to each other, such as DeCLUTR \cite{giorgi-etal-2021-declutr}, SimCSE \cite{gao-etal-2021-simcse}, TransEncoder \cite{liu2021trans} and DiffCSE \cite{chuang2022diffcse}. 
Another approach relies on autoencoders to compress a sentence into a latent vector representation and then reconstruct the original sentence, such as VGVAE \cite{chen-etal-2019-multi} and TSDAE \cite{wang-etal-2021-tsdae-using}.

Most unsupervised methods are based on textual sentences. In speech, current studies tend to center on acoustic word embeddings \cite[e.g.,][]{kamper2016deep,settle2016discriminative,Settle2017,Holzenberger2018,kamper2019truly}. Despite the progress, learning sentence-level embeddings for speechstill remains under-explored. In SUPERB benchmark for evaluating speech representations \citep{yang21c_interspeech}, spoken sentence similarity ranking is not yet listed as a downstream task. In recent works, it has been shown that spoken sentence semantic similarities can be learned via the visually grounded speech models \cite{merkx21_interspeech}. Multilingual spoken sentence embeddings can also be learned by using supervised multilingual text models as teacher models \cite{duquenne2021multimodal,khurana2022samu}. These methods more or less relied on labeled data such as speech-image pairs or multilingual sentence pairs. However, we propose unsupervised methods to induce semantic embeddings from speech signals only, and our methods can also utilize textual transcriptions to improve performance if they are available.

\section{Method}
\paragraph{Task formulation} The current task is to encode spoken utterances into low dimensional dense vectors such that semantically similar utterances are close to each other in the learned latent space. Given a speech signal $\boldsymbol{x}\in \mathbb{R}^{1\times N}=[x_1,x_2,\dots,x_N]$, our goal is to learn a neural network function $f_{enc}$ that converts  $\boldsymbol{x}$ to a fixed-dimensional vector $\boldsymbol{z}\in \mathbb{R}^d=f_{enc}(\boldsymbol{x})$, such that $\boldsymbol{z}$ encodes the semantic content of the original signal $\boldsymbol{x}$. For a certain semantically similar pair \{$\boldsymbol{z},\boldsymbol{z}^+$\} and a semantically dissimilar pair \{$\boldsymbol{z},\boldsymbol{z}^-$\} (as determined by human raters), it is expected that $sim(\boldsymbol{z},\boldsymbol{z}^+)>sim(\boldsymbol{z},\boldsymbol{z}^-)$, where $sim()$ is a similarity scoring function. 

It is further assumed that some forms of transcriptions of the original speech signal exist. Usually, a transcription of $\boldsymbol{x}$ take the form of a textual sequence $\boldsymbol{y}\in\mathbb{R}^{1\times M}=[y_1,y_2,\dots,y_M],N>M$. Such data are sometimes available as most speech datasets for ASR and TTS are organized as pairs of speech and texts.
However, in most scenarios the textual transcriptions are not available or too costly to create. In these cases, the transcriptions can be in the form of pseudo-units $\boldsymbol{\hat{y}}\in\mathbb{R}^{1\times L}=[\hat{y}_1,\hat{y}_2,\dots,\hat{y}_L],N>L$, which could be generated by an unsupervised  system for acoustic unit discovery. During training, these transcripts are used as the targets for the proxy tasks. However, in inference, the pretrained model can directly project speech into semantic embeddings. 

\subsection{Discretizing speech signals}
Acoustic unit discovery refers to the task of segmenting speech signals into discrete word-like or phone-like units \cite[e.g.,][]{lee-glass-2012-nonparametric,lee-etal-2015-unsupervised,ondel2016variational,kamper2019truly,Niekerk2020}. Annotating speech signals can sometimes be prohibitively costly for many languages and application domains. Unsupervised discovery of acoustic units can be used as a proxy of transcriptions to train speech systems, if the discovered acoustic units are consistent representations of speech. In our approach, acoustic units are treated as `pseudo-texts" to bootstrap the learning of semantic representations. 

We used pretrained speech transformer, HuBERT, to discretize speech signals into `hidden units', which were proposed in \citet{baevski2021unsupervised} and \citet{lakhotia-etal-2021-generative}. After passing speech signals into HuBERT, the hidden states of the sixth layer were extracted and a k-means clustering algorithm was applied on the hidden states to quantize them into discrete clusters. The sequence of cluster indexes, after deduplication by merging consecutive same indexes, are the hidden units representing the original speech (see Figure~\ref{fig:wavembed_archi}). The discrete hidden units remove certain paralinguistic and non-linguistic variations such as speaker voice traits and background noises, so they can be considered a normalized representations of the speech content (though many phonetic variations are still present) \citep{lee2021textless}.

We used the \texttt{textless-lib} \citep{kharitonov2022textless} to convert speech signals into discrete hidden units. 
We selected \texttt{hubert-base-ls960} as the base speech encoder and set the number of clusters to 50, 100 and 200. 
After speech were discretized into sequences of hidden units, sentence-piece tokenizers \cite{kudo-richardson-2018-sentencepiece} were trained on them to shorten the sequence length (see Appendix~\ref{app: tokenizers}). There is evidence showing that re-tokenzing hidden units are generally beneficial for language modeling and downstream tasks \cite{ren2022speech,wu2022wav2seq}.

\subsection{S-HuBERT}

The first approach, S-HuBERT, is to transfer the knowledge of a well-learned text embedding model to a speech embedding model \cite{duquenne2021multimodal,khurana2022samu}, in which pretrained supervised textual embeddings are adopted as the teacher models and speech models are trained to align with the text embeddings in the same latent space. 

Here we also extend this approach to the unsupervised learning domain. The proposed utilizes an unsupervised sentence embedding model with transcriptions, and then transfers the knowledge of a textual sentence embedding model to an acoustic sentence embedding model (S-HuBERT) by leveraging the correspondence between speech and its transcriptions. In the absence of textual transcriptions, the hidden units can be processed as pseudo-texts to induce unsupervised meaning embeddings.

We mainly investigate two approaches to train unsupervised (pseudo-)text embedding models, namely, SimCSE \citep{gao-etal-2021-simcse} and TSDAE \citep{wang-etal-2021-tsdae-using}. If these two types of models are trained with hidden units, they are referred to as Hu-SimCSE and Hu-TSDAE respectively, in order to distinguish them from the text-based models. 

\paragraph{SimCSE} The unsupervised SimCSE \cite{gao-etal-2021-simcse} is a contrastive learning framework for textual sentence embeddings. It takes a sentence as input and uses the same sentence as the target with only the dropout noises. As pretrained transformers such as BERT and RoBERTa apply a dropout mask of 10\%, the same sentence will result in slightly different hidden states in multiple passes and can be treated as positive pairs in contrastive learning. We trained SimCSE models to induce sentence meaning from text transcripts before transferring the knowledge to a speech model.
For modeling hidden units, we first pretrained a BERT model on hidden units, which were converted from the whole speech corpus. Then Hu-SimCSE was initiated with the pretrained hidden-unit BERT for training.

\paragraph{TSDAE}
Transformer-based Sequential Denoising AutoEncoder (TSDAE) is a denoising encoder-decoder model that encodes a corrupted text sequence into a dense vector and decodes the original text sequence. We trained text-based TSDAE models following as closely as possible the settings specified by \citet{wang-etal-2021-tsdae-using}.  However, slightly different hyperparameters were adopted for Hu-TSDAE. In the original TSDAE, tokens in the input sequence are randomly deleted with a ratio of 0.6. We found that deleting tokens in the input hidden units significantly hurt performance Instead, using the same uncorrupted sequence of hidden units as both inputs and targets achieved much better performance in our hyperparameter tuning experiments (see Appendix~\ref{app:deletion_rate}).

\paragraph{Language modeling on discrete units}
Both SimCSE and TSDAE models were intialized with pretrained transformer checkpoints. In addition to publicly available text-based pretrained mdoels, we also pretrained hidden-unit based pretrained transformers. Given a corpus of hidden units converted from raw speech, transformer-based language models were pretrained to learn the statistical regularities in sequences of hidden units. We adopted the same model architecture as BERT \cite{devlin-etal-2019-bert} (\texttt{bert-base-uncased}) and used the masked language modeling task with a masking rate of 15\%. However, the next sentence prediction task was discarded, because it was not found to significantly affect the model performance \cite{liu2019roberta,lan2019albert}.

\paragraph{Knowledge distillation} We transferred the knowledge from a pretrained textual sentence embedding model (SimCSE or TSDAE) into a speech embedding model through teacher-student training \cite{duquenne2021multimodal}. Here the teacher model was pretrained text embedding model whereas the student model was the pretrained speech model, HuBERT \cite{hsu2021hubert}.

We used contrastive learning for training S-HuBERT \cite{sun-etal-2020-contrastive,wu2021disco,ye2022cross}. Given a speech embedding $\boldsymbol{z}_i$ and its corresponding (pseudo-)text embedding $\boldsymbol{\widetilde{z}}_i^+$ with in-batch negative samples, the InfoNCE loss \cite{oord2018representation} is computed as
\begin{equation}
    \mathcal{L}_{\text{infoNCE}} = -\log\frac{e^{sim(\boldsymbol{z}_i,\boldsymbol{\widetilde{z}}_i^+)/\tau}}{\sum_{j=1}^N e^{sim(\boldsymbol{z}_i,\boldsymbol{\widetilde{z}}_j^+)/\tau}}
\end{equation}
where $\tau$ is the temperature parameter and $sim()$ is the cosine similarity function $sim(\boldsymbol{z}_1,\boldsymbol{z}_2)=\boldsymbol{z}_1^T\boldsymbol{z}_2/||\boldsymbol{z}_1||\cdot||\boldsymbol{z}_2||$. $\tau$ was set to 0.05 in all experiments. In order to keep a large number of negative samples, we maintained a dynamic memory bank of negative samples \cite{he2020momentum}. In each iteration, textual representations in the last mini-batch are enqueued into the memory bank, whereas the oldest textual representations in the bank are dequeued. The text model is frozen throughout training. A comparison of InfoNCE and MSE loss is available at Table~\ref{tab:pooling-and-loss} in Appendix~\ref{app:pooling-and-loss}.
 
\subsection{WavEmbed}

The WavEmbed is a sequential autoencoder \cite{10.5555/1756006.1953039,hill-etal-2016-learning,wang-etal-2021-tsdae-using}, which encodes a speech signal $\boldsymbol{x}$ into a fixed-dimensional vector $\boldsymbol{z}$ and decodes the vector representation $\boldsymbol{z}$ using only the encoded vector. The vector $\boldsymbol{z}$ is used as the semantic representation.
The decoded discrete representations can be actual texts $\boldsymbol{y}$ or sequences of hidden acoustic units $\boldsymbol{\hat{y}}$.  The proposed method is inspired by the TSDAE \cite{wang-etal-2021-tsdae-using}, which learns effective unsupervised sentence embeddings through a denoising encoder-decoder model that encodes a corrupted text sequence into a dense vector and decodes the uncorrupted one. WavEmbed generalizes the original TSDAE to acoustic signals and can learn semantic representations of speech through reconstructing not only the texts but also the hidden acoustic units discovered unsupervisedly. 

Yet WavEmbed differs from TSDAE in some aspects. TSDAE's encoder and decoder components are all text-based, whereas WavEmbed utilizes a speech encoder. TSDAE relies on the denoising reconstruction as a proxy task, in which the model is trained to recover the original sentence from the embedding of the corrupted sentence (word deletion with a ratio of 0.6). However, WavEmbed reconstructs a discrete sequence from the embedding of a corresponding spoken sentence but no corruptions except the standard dropout is applied to the speech signal. In addition, WavEmbed uses self-attention pooling to pool the encoder hidden states rather than the average pooling in TSDAE, as self-attention pooling is more effective than mean or max pooling for sentence-level speech emebddings \cite[see][and Table~\ref{tab:pooling-and-loss} in Appendix~\ref{app:pooling-and-loss}]{khurana2022samu}. 

The encoder $f_{enc}$ consists of two parts, a pretrained speech transformer $f_{\boldsymbol{S}}$ for speech feature extraction and a self-attention pooling layer for pooling. Let $\boldsymbol{H}\in\mathbb{R}^{T\times d}=[\boldsymbol{h}_1,\boldsymbol{h}_2,\dots,\boldsymbol{h}_T]$ be the hidden states of a speech transformer model $f_{\boldsymbol{S}}$ given a speech signal $\boldsymbol{x}$. The self-attention pooling operation \cite{Safari2020} can be computed as:
\begin{equation}
    \boldsymbol{H} = f_{\boldsymbol{S}}(\boldsymbol{x})
\end{equation}
\vspace{-0.2in}
\begin{equation}
    \boldsymbol{z} = \text{Softmax}(\boldsymbol{W}\boldsymbol{H}^T)\boldsymbol{H}
\end{equation}
where $\boldsymbol{W}\in\mathbb{R}^d$ is a learnable parameter during training. Given a semantic representation $\boldsymbol{z}$ of speech signal $\boldsymbol{x}$, the autoregressive decoder $f_{dec}$ predicts the hidden units $\boldsymbol{\hat{y}}$ that correspond to the content of the speech signal $\boldsymbol{x}$. 
\begin{equation}
    \boldsymbol{\hat{y}} = f_{dec}(\boldsymbol{z})
\end{equation}
The encoder-decoder model is trained with the standard negative log likelihood loss. 
\begin{equation}
    \begin{aligned}
    \mathcal{L} =& -\sum_1^L \log P(\hat{y}_l|\boldsymbol{z},\hat{y}_{l-1},\dots,\hat{y}_1) \\
    =& -\sum_1^L \log P(\hat{y}_l|f_{enc}(\boldsymbol{x}),\hat{y}_{l-1},\dots,\hat{y}_1)
    \end{aligned}
\end{equation}
The WavEmbed is trained to predict discrete acoustic units $\boldsymbol{\hat{y}}$ based on the speech signals $\boldsymbol{x}$. However, when textual transcripts for speech signals are available, the prediction targets can also be replaced with textual sequences $\boldsymbol{y}$ to enhance the learning of semantic content. Once the model is trained, the decoder is discarded, leaving only the speech encoder for extracting the semantic embeddings.

\section{Data}
\paragraph{Training data}
We used the trianing set of the 1700-hour English subset of Common Voice \cite{ardila-etal-2020-common} for training all speech encoders, which can be assessed through the HuggingFace \texttt{Datasets} library \cite{lhoest-etal-2021-datasets}. Speech signals were all downsampled from 44.1kHz to 16kHz to be compatible with HuBERT. To avoid memory issues, we limited the maximal length of speech to be 10 seconds, which only excluded less than 1\% of data.

\paragraph{Common Voice spoken sentence similarity}
Given that there is few evaluation data for spoken sentence similarities, we created the Common Voice Spoken Sentence Similarity (CVS) dataset based on the test set of the English subset of Common Voice, which contains a wide range of accents. Following the criteria of annotating STS test data \cite{agirre-etal-2016-semeval}, four proficient English users independently rated the similarity of 1149 sentence pairs on a scale of 0 to 5 (not similar to most similar). The average Spearman's rank correlation between the four raters reached 0.937.
The annotated data was randomly partitioned into the  development set (40\% , 459 sentence pairs) and the test set (60\%, 690 sentence pairs). Score distributions and additional details are provided in Appendix~\ref{app:cvss}.

\begin{table*}[t]
\begin{adjustbox}{width=0.95\textwidth,center}
\begin{tabular}{lrrrrrrrr}
\toprule
\multicolumn{1}{c}{\multirow{2}{*}{\bf Models}} & \multicolumn{5}{c}{\bf Synthetic speech}  & \multicolumn{3}{c}{\bf Natural Speech} \\  \cmidrule(lr){2-6} \cmidrule(lr){7-9} 
 & \textbf{STS12} & \textbf{STS13} & \textbf{STS14} & \textbf{STS15} & \textbf{STS16} & \textbf{STS}  & \textbf{CVS-dev}  & \textbf{CVS-test}  \\ \toprule
\multicolumn{8}{l}{\it \textbf{A.} Unsupervised textual sentence embeddings (evaluated on texts)} \\\midrule
\textit{\textcolor{NavyBlue}{TSDAE}}  &  \textbf{\textit{\textcolor{NavyBlue}{55.2}}} & \textbf{\textit{\textcolor{NavyBlue}{67.4}}} & \textbf{\textit{\textcolor{NavyBlue}{62.4}}} & \textbf{\textit{\textcolor{NavyBlue}{74.3}}} & \textbf{\textit{\textcolor{NavyBlue}{73.0}}} & - & - & - \\
TSDAE-CV-text & 51.3  & 67.3 & 60.6 & 71.5 & 74.9 & \textbf{65.5}  & \textbf{90.7} & \textbf{89.5}\\ 
TSDAE-ASR-text & 50.8  & 67.9 & 58.5 & 70.2  & 74.6 &64.6 & 90.0  & 88.5\\\midrule
\multicolumn{8}{l}{\it \textbf{B.} Unsupervised spoken sentence embeddings (trained with texts)} \\\midrule
WavEmbed (BERT decoder)         &  48.8     &   53.8    &   52.7    &  66.1     &  63.6     &  55.4     &   \textbf{78.4}          &     78.7         \\ 
WavEmbed (RoBERTa decoder)         &  \textbf{52.1}     &  \textbf{56.1}     & \textbf{54.7}      &   \textbf{68.4}    &  \textbf{66.1}     &   \textbf{58.6}    &      78.1       & \textbf{79.3}             \\ \midrule
\multicolumn{8}{l}{\it \textbf{C.} Unsupervised spoken sentence embeddings (trained with ASR transcripts)} \\\midrule
WavEmbed (BERT decoder)   & \textbf{52.1} & \textbf{55.9} & \textbf{55.6} & \textbf{68.0} & \textbf{66.2} & \textbf{57.9} & 77.5 & \textbf{79.5}           \\ 
WavEmbed (RoBERTa decoder)  & 49.3 & 53.0 & 52.2 & 64.9 & 64.0 & 54.6 & \textbf{77.9} & 78.1         \\ \midrule
\multicolumn{8}{l}{\it \textbf{D.} Unsupervised spoken sentence embeddings (trained with hidden units)} \\\midrule
HuBERT - Avg. Last hidden states & \textbf{47.3} & 49.5 & 44.1 & 58.3 & 49.2 & 33.2 & 1.5 & 4.4  \\ \midrule 
WavEmbed - 50 C         &  44.0 & 51.5 & 47.6 & 62.7 & 56.3 & 49.8 & 62.6 & 59.8      \\ 
WavEmbed - 100 C        &  45.3 & \textbf{53.4} & \textbf{50.3} & 63.3 & \textbf{59.7} & \textbf{50.0} & 65.6 & \textbf{61.2}      \\ 
WavEmbed - 200 C         &  43.8 & 49.9 & 48.7 & \textbf{63.5} & 58.4 & 47.4 & 64.8 & 59.2  \\  \midrule
WavEmbed - 100 C - 1000 V   & 45.2 & 49.7 & 47.1 & 61.8 & 58.1 &  45.7 & 67.8 & 60.9     \\ 
WavEmbed - 100 C - 3000 V & 44.3  & 50.1 & 47.1 & 61.4 & 57.3 & 45.5  & 68.4 & 61.0 \\ 
WavEmbed - 100 C - 5000 V   & 43.8 & 48.9 & 45.9 & 60.9 & 57.5 & 44.8 & 67.9 & 61.0   \\ 
WavEmbed - 100 C - 8000 V    & 43.1 & 48.8 & 45.7 & 60.3 & 58.8 & 43.5 & \textbf{68.9} & 61.1    \\ 
WavEmbed - 100 C - 12000 V     &  43.3 & 47.6 & 44.4 & 60.3 & 57.5 & 42.8 & 66.0 & 60.2
 \\ \bottomrule
\end{tabular}
\end{adjustbox}
\caption{Evaluations based on the Spearman's rank correlation ($\times100$). The italicized blue fonts suggest that results were directly taken from \citet{wang-etal-2021-tsdae-using}. The bold numbers indicate the best performance within each subsection. Note. C $\rightarrow$ number of clusters; V $\rightarrow$ size of vocabulary; CV-text $\rightarrow$ Common Voice texts. } 
\label{tab:wavembed}
\end{table*}

\paragraph{Spoken STS}
Additionally, we used the Spoken STS data collected by \citet{merkx21_interspeech} for evaluation, which is publicly available online\footnote{\url{https://easy.dans.knaw.nl/ui/datasets/id/easy-dataset:237533}}. The Spoken STS include synthetic and natural recordings of sentences in the Semantic Textual Similarity (STS) dataset, covering benchmarks from STS12 to STS16 \cite{agirre-etal-2012-semeval,agirre-etal-2013-sem,agirre-etal-2014-semeval,agirre-etal-2015-semeval,agirre-etal-2016-semeval}. The STS datasets include paired sentences with human labelled similarity scores. The synthetic speech data contain all sentence pairs in STS datasets, which were synthesized via Google's Wavenet using six voices (three males and three females) with a US accent. The natural speech data contains 638 pairs of randomly selected sentences evenly distributed across the STS datasets. These sentences were produced by four speaker (2 females and 2 males) with a North American accent. The synthetic and natural speech was only used as the test set. None of our models had seen any STS sentences during training. During evaluation, for each sentence pair, we averaged the similarity scores for all possible combinations of speakers to derive the final score.

\begin{table*}[th]
\begin{adjustbox}{width=0.95\textwidth,center}
\begin{tabular}{lrrrrrrrr}
\toprule
\multicolumn{1}{c}{\multirow{2}{*}{\bf Models}} & \multicolumn{5}{c}{\bf Synthetic speech}  & \multicolumn{3}{c}{\bf Natural Speech} \\  \cmidrule(lr){2-6} \cmidrule(lr){7-9} 
 & \textbf{STS12} & \textbf{STS13} & \textbf{STS14} & \textbf{STS15} & \textbf{STS16} & \textbf{STS}  & \textbf{CVS-dev}  & \textbf{CVS-test}  \\ \toprule
 \multicolumn{8}{l}{\it \textbf{A.} Supervised text embeddings}
 \\\midrule
\textit{\textcolor{NavyBlue}{SimCSE-sup-BERT}}   & \textit{\textcolor{NavyBlue}{75.3}} &\textit{\textcolor{NavyBlue}{84.7}} &\textit{\textcolor{NavyBlue}{80.2}} &\textit{\textcolor{NavyBlue}{85.4}} &\textit{\textcolor{NavyBlue}{80.8}}    &  \textbf{81.4}  &  \textbf{92.9} &  \textbf{92.3} \\
\textit{\textcolor{NavyBlue}{SimCSE-sup-RoBERTa}}  &  \textbf{\textit{\textcolor{NavyBlue}{76.5}}} &  \textbf{\textit{\textcolor{NavyBlue}{85.2}}} &  \textbf{\textit{\textcolor{NavyBlue}{81.0}}}& \textbf{\textit{\textcolor{NavyBlue}{86.0}}}& \textbf{\textit{\textcolor{NavyBlue}{82.6}}}    & 80.2 & 91.2 & 90.9       \\ \midrule

\multicolumn{8}{l}{\it \textbf{B.} Unsupervised text embeddings} \\\midrule
\textit{\textcolor{NavyBlue}{SimCSE-unsup-BERT}} &\textit{\textcolor{NavyBlue}{68.4}}&\textit{\textcolor{NavyBlue}{82.4}}&\textit{\textcolor{NavyBlue}{74.4}}&\textit{\textcolor{NavyBlue}{80.9}}&\textit{\textcolor{NavyBlue}{78.6}}  &77.6  & 89.6 & \textbf{89.0}\\
\textit{\textcolor{NavyBlue}{SimCSE-unsup-RoBERTa}} &  \textbf{\textit{\textcolor{NavyBlue}{70.2}}} &  \textbf{\textit{\textcolor{NavyBlue}{81.8}}} &  \textbf{\textit{\textcolor{NavyBlue}{73.2}}} &  \textbf{\textit{\textcolor{NavyBlue}{81.4}}} &  \textbf{\textit{\textcolor{NavyBlue}{80.7}}}   &  \textbf{77.7} & 88.0 & 87.3  \\ 
SimCSE-unsup-BERT-CV-text & 61.2  & 77.9 & 69.9 & 77.6 &76.8  & 73.1 &  \textbf{89.7} &  87.7\\
SimCSE-unsup-BERT-ASR-text &57.7 & 69.0  & 64.2  & 74.5 & 68.7 & 66.5 & 85.5 & 86.1\\\midrule
\multicolumn{8}{l}{\it \textbf{C.} Supervised text embeddings $\rightarrow$ spoken sentence embeddings}
 \\\midrule
S-HuBERT (\textit{\textcolor{NavyBlue}{SimCSE-sup-BERT}})   & 57.3 & 63.9 & 62.6 & 72.6 & 67.6 & 64.5 & 81.1 & 79.7      \\
S-HuBERT (\textit{\textcolor{NavyBlue}{SimCSE-sup-RoBERTa}})   &   \textbf{61.5}   &    \textbf{68.0}    &    \textbf{65.1}    &   \textbf{75.2}     &   \textbf{70.8}     &    \textbf{69.0}    &    \textbf{81.3} &   \textbf{81.9}            \\ \midrule
\multicolumn{8}{l}{\it \textbf{D.} Unsupervised text embeddings $\rightarrow$ spoken sentence embeddings} \\\midrule
S-HuBERT (\textit{\textcolor{NavyBlue}{SimCSE-unsup-BERT}})  & 53.0 & 59.6 & 58.5 & 67.8 & 65.4 & 60.5 &  \textbf{80.0} & 80.8   \\
S-HuBERT (\textit{\textcolor{NavyBlue}{SimCSE-unsup-RoBERTa}})  &   \textbf{60.8}     &    \textbf{70.8}    &    \textbf{65.9}    &  \textbf{76.5}      &   \textbf{72.8}     &     \textbf{70.1}   &     79.8        &     \textbf{81.6}         \\ 
S-HuBERT (TSDAE-CV-text) & 38.6 & 43.6 & 45.6 & 55.4 & 54.6 & 45.4 & 77.8 & 76.2 \\
S-HuBERT (TSDAE-ASR-text) & 43.1 & 47.0 & 48.0 & 58.7 & 55.9 & 48.0 & 79.5 & 75.6 \\\midrule
\multicolumn{8}{l}{\it \textbf{E.} Unsupervised hidden-unit embeddings $\rightarrow$ spoken sentence embeddings} \\\midrule
S-HuBERT (Hu-SimCSE-100 C-8000 V)    & 37.4 & 36.8 & 32.3 & 44.5 & 39.6 & 26.6 & 35.2 &   34.6   \\ 
S-HuBERT (Hu-TSDAE-100 C-1000 V)   & \textbf{47.8} &  \textbf{53.8} &  \textbf{51.6} &  \textbf{63.8} &  \textbf{60.4} &  \textbf{50.3} &  \textbf{61.5} & \textbf{55.3}     \\ \bottomrule
\end{tabular}
\end{adjustbox}
\caption{Evaluations based on the Spearman's rank correlation ($\times100$). The italicized blue font suggests that results or checkpoints were directly taken from \citet{gao-etal-2021-simcse}. Models in parentheses are teacher models whereas models outside parentheses are student models. The bold numbers indicate the best performance within each subsection. Note. C $\rightarrow$ number of clusters; V $\rightarrow$ vocabulary size; CV-text $\rightarrow$ Common Voice texts.  }
\label{tab:distil}
\end{table*}

\section{Experiments}

Three different types of transcriptions were compared in this study, namely, (ground truth) text transcripts, text transcripts recognized by an ASR model, and hidden units. ASR transcripts were additionally considered because for some languages ASR systems can be used to assist the development of speech retrieval. We used the \texttt{hubert-large-ls960-ft}  \cite{hsu2021hubert} via HuggingFace Hub to transcribe the Common Voice English subset to texts.

For WavEmbed models, we trained multiple variants based on texts, ASR transcriptions and hidden units. The audio encoder was initiated with the \texttt{hubert-base-ls960} checkpoint. The decoder was also initiated with pretrained weights according to their outputs, including \texttt{bert-base-uncased}, \texttt{gpt2-base}, \texttt{roberta-base}, as well as pretrained transformers for hidden units on different vocabulary sizes.

For S-HuBERT, we first trained SimCSE and TSDAE on Common Voice texts, ASR transcriptions and hidden units, all of which were also initiated with pretrained weights according to respective inputs. Specifically for SimCSE, additionally four existing model checkpoints from \citet{gao-etal-2021-simcse} were adopted without any training nor fine-tuning. Then the HuBERT base model with a self-attention pooling layer and a projection head on top was finetuned to aligned with the semantic embeddings of the text models in the same latent space. For Hu-SimCSE and Hu-TSDAE, we only selected the best performing models to perform model distillation.

Given that there are several different models implemented here, detailed hyperparameter settings are given at Appendix~\ref{app: hyperparameters}.
The implementation of transformers was modified based on the \texttt{transformers} library \cite{wolf-etal-2020-transformers} and the \texttt{sentence\_transformer} library \cite{reimers-gurevych-2019-sentence}. All our models were trained on a single A40 GPU with 48GB of memory. Unless otherwise stated, all models were initialized with pretrained models, as pretrained weights were found to improve downstream performance \citep[see][and Table~\ref{tab:weights} in Appendix~\ref{app:weight}]{rothe-etal-2020-leveraging}.

\section{Results and discussions}
In general, we show that semantic representations can be learned directly from acoustic signals through the proxy of hidden units. Here we summarize the main results and the ablation analyses (see Appendix~\ref{app:ablation} for additional results).

\paragraph{The self-reconstruction approach is more effective in modeling hidden units than contrastive learning.}  In general, the WavEmbed method is far more effective than S-HuBERT when transcriptions are not available. The WavEmbed models with hidden-unit targets all achieve moderate correlation with human judgments (Table~\ref{tab:wavembed} D), as compared to the weak correlations with humans for the S-HuBERT from Hu-SimCSE (Table~\ref{tab:distil} E). Even for directly modeling on hidden units without audio, the self-reconstruction based Hu-TSDAE far outperform the contrastive learning based Hu-SimCSE in terms of Spearman correlations. \citet{gao-etal-2021-simcse} noted that SimCSE depends heavily on a good pretrained LM. Yet our hidden unit-based BERT is trained on a relatively small corpus without a sentence-level task, which presumably is the reason why the performance of Hu-SimCSE is sub-optimal.   

\paragraph{Re-tokenizing hidden units does not always help.} \citet{ren2022speech} and \citet{wu2022wav2seq} found that re-tokenizing hidden units can improve the performance of downstream tasks, yet our findings here are inconclusive. For WavEmbed (Table~\ref{tab:wavembed} D), re-tokenizing hidden units does not improve model performance. However, for Hu-SimCSE and Hu-TSDAE, the results are almost contrary. Using sentence piece units learned from the raw hidden units tends to improve Spearman correlation in these hidden unit-based models (see Table~\ref{tab:tsdae-hu-impact} and~\ref{tab:simcse-hu-impact} in Appendix~\ref{app:hu_impact}).

\paragraph{Supervised training with texts is the most effective method.} This holds even if supervised data are in another modality. In Table~\ref{tab:distil} C, it is apparent that transferring weights of supervised embedding models to S-HuBERT achieves the best evaluation performance across all evaluation benchmarks. This result may suggest that, given well trained supervised text models, the need for created labeled speech data might not be very critical.

\paragraph{Training with text transcripts are better than with pseudo-units.}  In both Table~\ref{tab:wavembed} and~\ref{tab:distil}, a comparison between text-based and hidden unit-based models clearly shows that text-based WavEmbed or S-HuBERT outperformed their hidden unit-based counterparts by a very large margin (ranging from 5$\sim$40 in absolute points). Compared with texts, hidden units are noisy representations of the speech content, such that speaker identities were still detectable from hidden units converted from speech \cite{kharitonov2022textless}. Such content-irrelevant information introduces many noises into model training, therefore leading to sub-optimal performance. WavEmbed and Hu-TSDAE are more robust to local noises, as they force the speech embedding to capture the global property of the sentence through self-reconstruction \cite{wang-etal-2021-tsdae-using}. 
\begin{figure}[tbh]
    \centering
    \includegraphics[width=\linewidth]{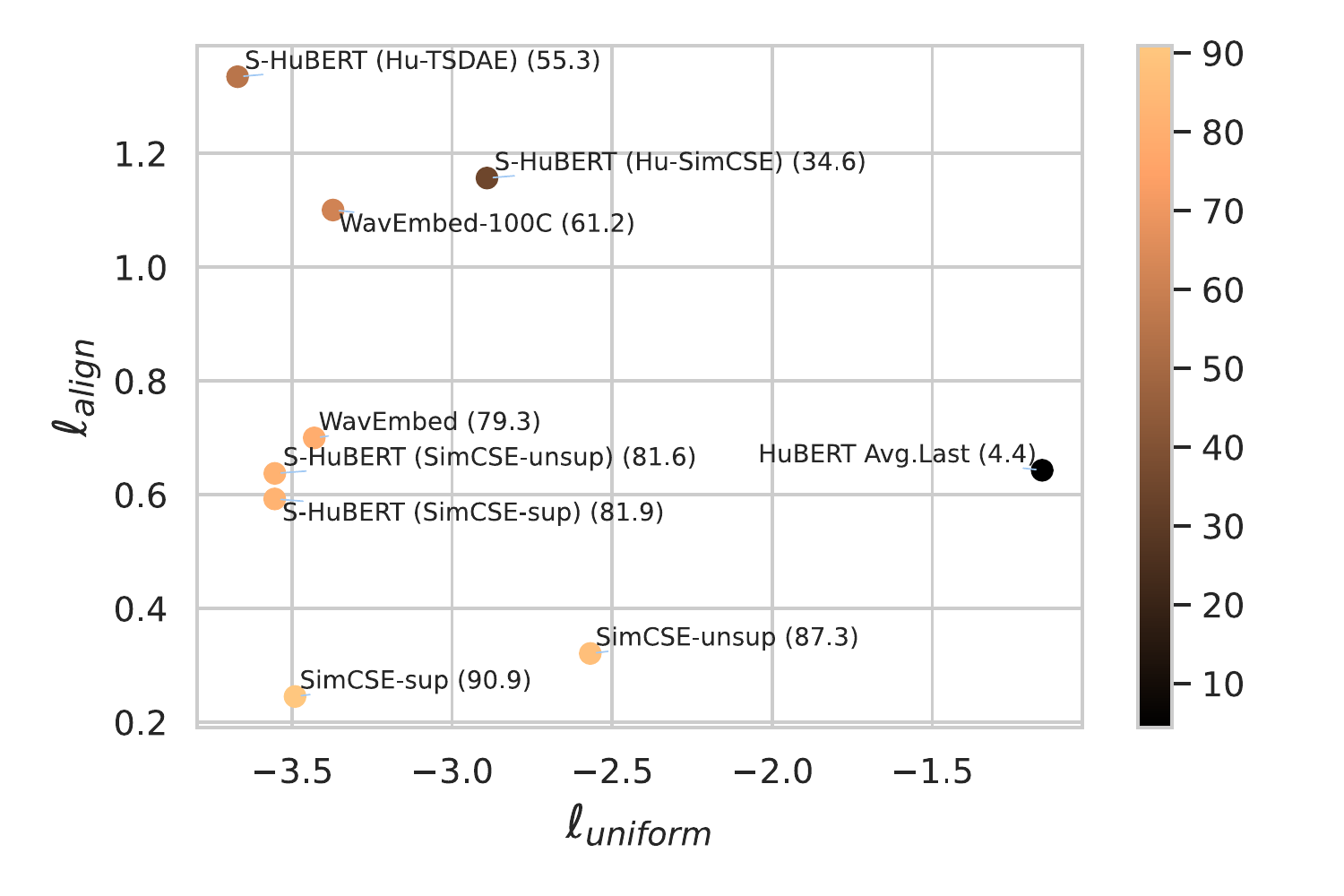}
    \caption{$\ell_{align}-\ell_{uniform}$ plot. All SimCSE models and the WavEmbed decoder were based on RoBERTa-base. The bracketed number indicates Spearman correlation ($\times$100). Low $\ell_{align}$ \textbf{and} low $\ell_{uniform}$ are desirable qualities of embeddings. }
    \label{fig:align-uniform}
\end{figure}

In order to measure the quality of the learned embeddings, we compute the metrics for alignment and uniformity proposed by \citet{wang2020understanding}, which are as follows.
\begin{equation}\label{eq:align}
    \ell_{align} \triangleq \mathop{\mathbb{E}}_{x,x^+\sim P_{pos}}||f(x) - f(x^+)||^2
\end{equation}
\begin{equation}\label{eq:uniform}
    \ell_{uniform} \triangleq \log \mathop{\mathbb{E}}_{x,y \overset{\text{i.i.d}}{\sim} P_{data}} e^{-2||f(x)-f(y)||^2}
\end{equation}
where $x^+$ refers to a positive sample for $x$ and $P_{pos}$ is the set of positive samples. The analysis was based on the CVS test set. Sentence pairs with a score $\geq 4$ were treated as positive pairs for Equation~\ref{eq:align}. All possible combinations of sentence pairs were used to compute Equation~\ref{eq:uniform}. Figure~\ref{fig:align-uniform} shows the $\ell_{align}$-$\ell_{uniform}$ plot for different models \cite{wang2020understanding,gao-etal-2021-simcse}. Lower $\ell_{align}$ suggests that positive instances are closer to each other in the embedding space, whereas lower $\ell_{uniform}$ indicates that 
random representations are more uniformly scattered, both of which imply a good embedding space. Text-based SimCSE models have both low $\ell_{align}$ and low $\ell_{uniform}$. For speech based WavEmbed and S-HuBERT, they tend to have low $\ell_{uniform}$ but high $\ell_{align}$, so distances between positive samples are relatively far from each other. The average last hidden states of HuBERT has a very low $\ell_{uniform}$, suggesting that different embeddings are not well separated. It seems to perform well on STS (Table~\ref{tab:wavembed} D), because STS datasets were recorded from limited speakers and it might be using similarities of voice traits (see the first row of Table~\ref{fig:wavembed-corr}), which also explains why this method fails on the CVS dataset that contains a diversity of speakers . Generally speaking, for speech models not trained with texts, spoken utterances from the same speaker are always less separated than utterances from different speakers (see Figure~\ref{fig:wavembed-corr} and~\ref{fig:shubert-corr} in Appendix~\ref{app:vis-pred}).

\paragraph{ASR transcripts can be a good proxy for ground truth transcripts.} Modern ASR systems can achieve very low word error rate (WER) on certain high resource languages. Our experiments with recognized texts of Common Voice audio indicates that they are slightly inferior to the ground truth transcriptions in performance but the gap is pretty narrow (see Table~\ref{tab:wavembed}. B and C; Table~\ref{tab:distil}. B). In the absence of texts, it is possible to utilize ASR systems to quickly create training data for speech retrieval models. This option currently might only work well for a few high resource languages such as English, Chinese and Spanish. The recent advancement in unsupervised ASR \cite{baevski2021unsupervised,liu2022towards} or multilingual ASR \cite{li22aa_interspeech}, however, might make it possible to extend the proposed methods to more low resource languages.


\paragraph{WavEmbed and S-HuBERT are useful in different scenarios.} Our results have shown that WavEmbed works better when only speech signals are available, whereas S-HuBERT can leverage more powerful text models if transcripts are provided. Furthermore, since S-HuBERT aligns speech embeddings with text embeddings in a common semantic space, it is also capable of performing cross-modal semantic retrieval between speech and texts.  

\section{Conclusions and future directions}
In summary, we investigated two unsupervised approaches, WavEmbed and S-HuBERT, to induce semantic embeddings from speech signals, which are significantly correlated with human judgements of semantic similarities. We experimented these approaches with different settings and found that performance can be improved through leveraging transcribed texts, even when the transcriptions were from slightly inaccurate ASR systems. We believe that our study is an essential step towards direct speech-to-speech search and simplifying the pipeline of SLU. 

Currently, the proposed methods do not depend on any specific features of certain languages, and therefore should also work for other languages. However, it remains unclear how language specific properties can affect the actual performance. In the future, we plan to extend our methods to other languages than English. 

Generally speaking, spoken sentence similarity task is more challenging than its text counterparts, because speech has far more variability than texts, which can be caused by various physiological, psychological and social factors (e.g., vocal tract morphology, gender, accents and emotional states). It remains to be explored how these variabilities (some of which are related to expressing meaning) can be taken into account in computational models. Speech conveys information beyond its textual content. For example, prosody can reverse the meaning of a sentence from being non-ironic to ironic. While we only focus on the semantic similarity of textual content in speech, we also envision that spoken sentence similarity tasks in the future should include non-textual features such as prosody and emotion.

\section*{Acknowledgements}
We thank Professor David Jurgens at University of Michigan for his comments on the early versions of this manuscript. We are grateful for Danny Merkx for sharing the Spoken STS dataset. We would also like to thank three anonymous reviewers and the area chair for their suggestions, which helped improve this article greatly.

\section*{Limitations}
Our study is still limited in several ways. First, there is still a large gap between unsupervised learning from speech and unsupervised methods with texts. Part of the reason is that the performance is bounded by the relatively inaccurate acoustic unit discovery method that still retains many paralinguistic information such as voice traits \cite{lee2021textless}. One potential future direction to overcome this limitation is to develop better unsupervised acoustic unit discovery systems. 

Secondly, our proposed methods still require a large amount of data to pre-train and train, making it still difficult to extend this method to low resources. Large-scale pretrained models such as HuBERT, BERT and RoBERTa are only available for a limited number of languages that can offer huge data for pretraining. It is paramount to investigate data efficient methods that can work on low resource languages. Multilingual pretrained models such as XLS-R \cite{babu2021xls} might be a potential base model for low resource languages, but more work should be done. 

Given that transformers have a complexity of $\log(n^2)$, it might not be the optimal model for processing speech, which often have very long sequence length. In our experiments, we also notice that training and finetuning HuBERT on speech is very inefficient. An important future direction is to search for better model architectures that are optimized for processing long sequences of speech.

\bibliography{anthology,custom}
\bibliographystyle{acl_natbib}

\newpage
\appendix

\section{Sentence-piece Tokenizers}
\label{app: tokenizers}
All tokenizers for hidden units were trained with the open-source package \texttt{SentencePiece}\footnote{\url{https://github.com/google/sentencepiece}}. They were trained on the whole Common Voice English training set, which was transcribed into hidden units with cluster numbers of 50, 100, and 200. The vocabulary sizes were \{1000, 3000, 5000, 8000, 12000, 20000\} for each cluster number. We used the arguments in Table~\ref{app:token_hyper}, otherwise default arguments were kept.

\begin{table}[H]
\centering
\begin{tabular}{lr}
\toprule
\textbf{Arguments} & \textbf{Value} \\ \midrule
model\_type & BPE \\
hard\_vocab\_limit & true \\
split\_by\_whitespace & false \\
pad\_id& 0 \\
bos\_id & 1 \\
eos\_id & 2 \\
unk\_id & 3 \\ 
bos\_piece & [CLS] \\
eos\_piece & [SEP] \\ 
unk\_piece & [UNK] \\
pad\_piece & [PAD] \\
user\_defined\_symbols & [MASK]\\
\bottomrule
\end{tabular}
\caption{Arguments for training sentence-piece tokenizers on hidden units.}
\label{app:token_hyper}
\end{table}

\section{Common Voice Sentence Similarity}\label{app:cvss}
The Common Voice Spoken Sentence Similarity dataset was created based on the test set of the English subset of Common Voice. To get the similarity of every pair of sentences in the original test set, we used two pretrained SimCSE models, that is, \texttt{sup-simcse-roberta-large} and \texttt{sup-simcse-bert-large-uncased}\footnote{\url{https://github.com/princeton-nlp/SimCSE}}. 

The similarity scores were averaged across two models, and divided into ten equal intervals between 0 and 1. Then 120 sentence pairs were randomly sampled from each interval, resulted in 1200 sentence pairs. Some pairs were removed due to low quality, leaving 1149 pairs in the final data. Four proficient English users independently rated the sentence pairs based on the textual content on a scale from 0-5, where 0 implies completely irrelevant and 5 means completely the same. We tried to make this dataset similar to those in STS SemEval tasks and followd their annotation criteria. The instructions were the same as the Figure 2 in \citet{agirre-etal-2012-semeval} and sample rated sentences were from the Table 1 of \citet{agirre-etal-2016-semeval}. The final similarity scores were averaged across four human raters. The rated data was partitioned into the development set and the test set with a 40/60 split. The distributions of similarity scores in these two sets are provided in Figure~\ref{fig:hist_test}.

\begin{figure}
    \centering
    \includegraphics[width=0.8\linewidth, trim={1.5cm 0.8cm 6cm 1cm}, clip]{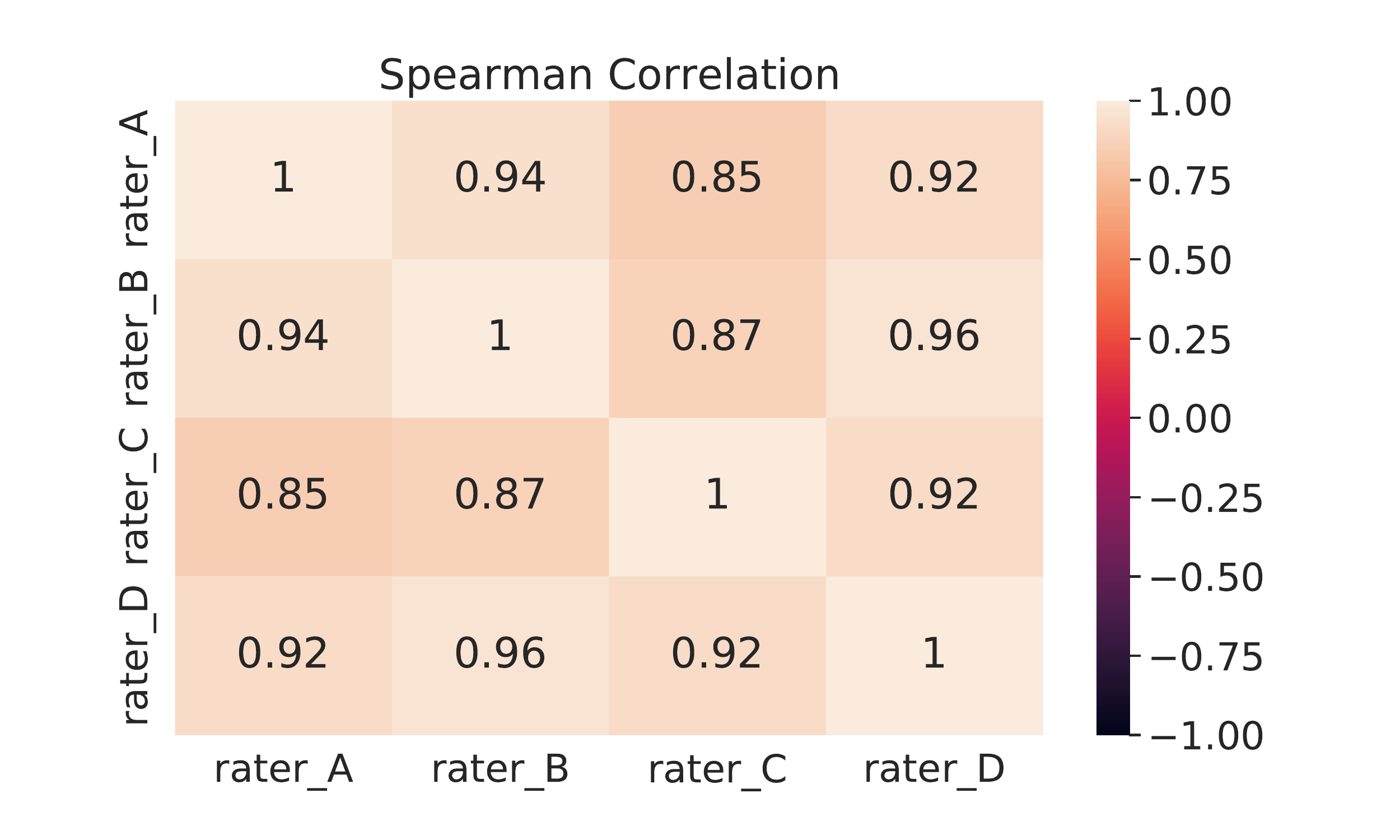}
    \caption{Pairwise Spearman's rank correlations between all four human raters. Sentence similarity ratings are highly consistent across human raters.}
    \label{fig:correlation}
\end{figure}

\begin{figure}
    \includegraphics[width=\linewidth, trim={0cm 0cm 2.5cm 0.5cm}, clip]{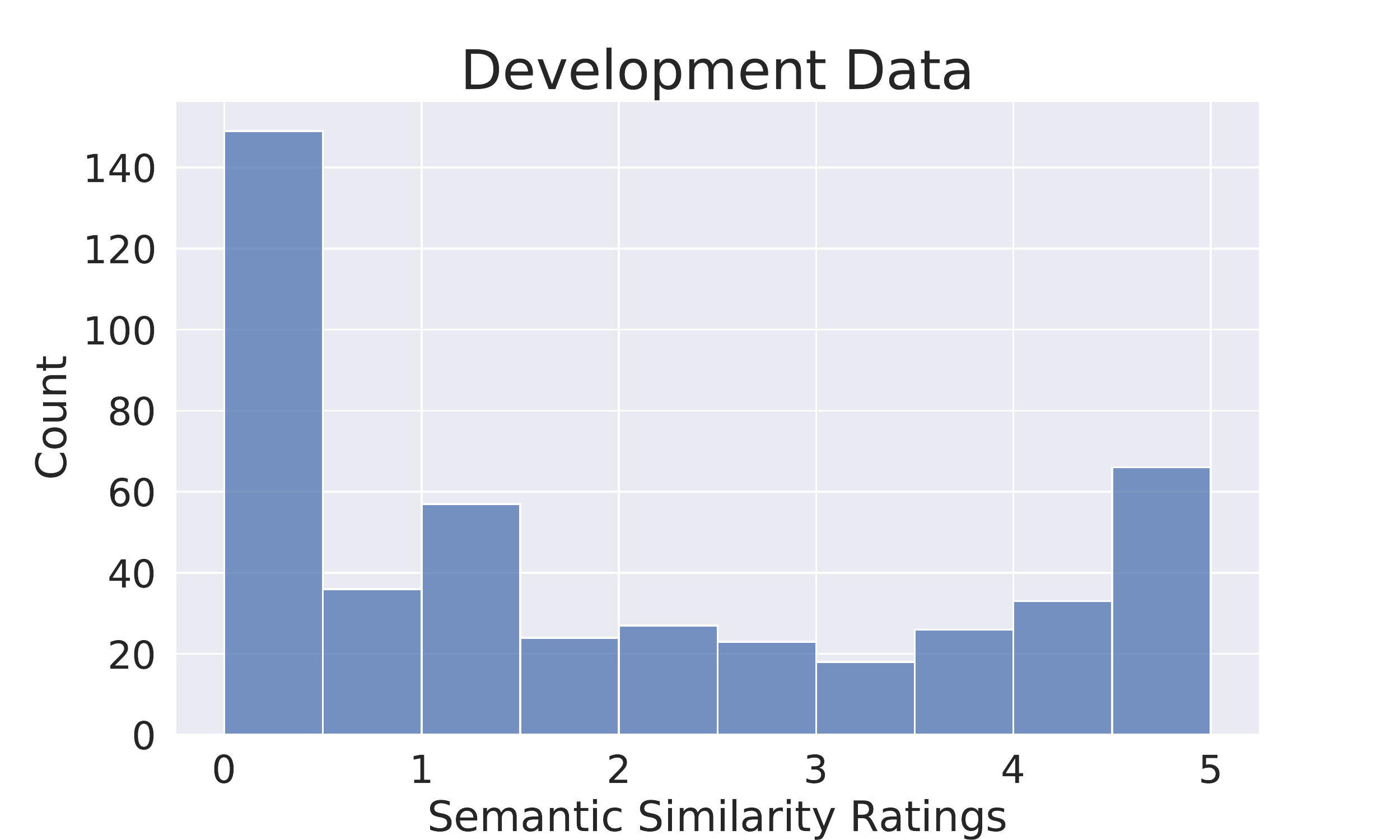}
    \includegraphics[width=\linewidth, trim={0cm 0cm 2.5cm 0.5cm}, clip]{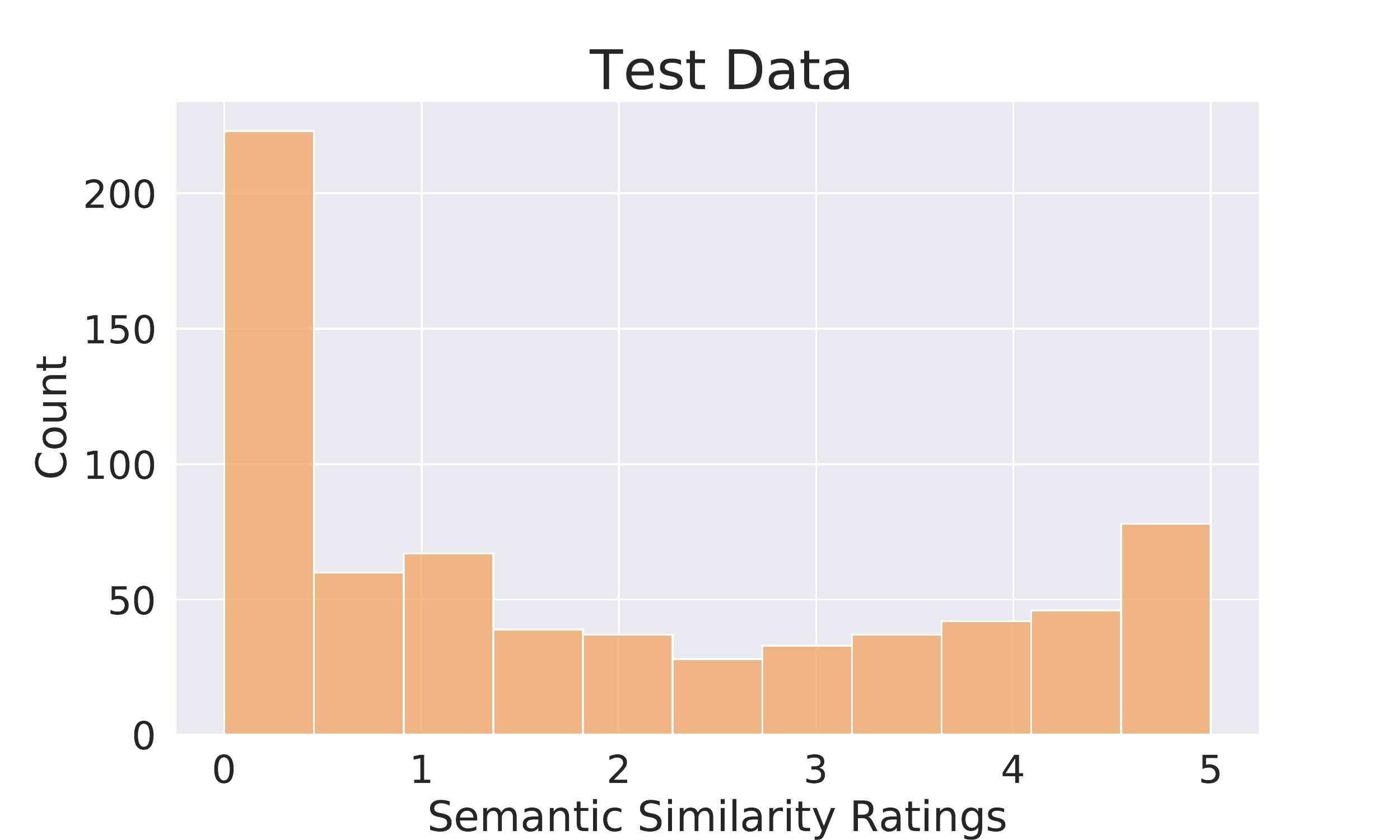}
    \caption{Distribution of ratings in Common Voice Spoken Sentence Similarity dataset.}
    \label{fig:hist_test}
\end{figure}

\section{Hyperparameters}
\label{app: hyperparameters}

\subsection{BERT on hidden units}
We trained 15 BERT models on hidden units from scratch. The detailed hyperparameters are given in Table~\ref{app:bert_hyper}, which were the same for all models. The number of clusters ranged from \{50, 100, 200\}. A BERT model was trained for each number of clusters. For hidden units of 100 and 200 clusters, they were further tokenized into vocabulary sizes of \{1000, 3000, 5000, 8000, 12000, 20000\} respectively. Then a BERT model was trained for each vocabulary size, resulting in 15 in total. Each took about 3.5$\sim$6 hours to train.

\begin{table}
\centering
\begin{tabular}{lr}
\toprule
\textbf{Hyparameters} & \textbf{Value} \\ \midrule
Epoch & 5 \\
Total update steps & $\sim$8k \\
Optimizer & AdamW \\
Initial learning rate & 1e-4 \\
Effective batch size & 512 \\
FP 16 & True \\
Masking probability & 15\% \\
Weight Initialization & random init. \\
Training task & MLM \\
Loss & Cross-entropy loss \\
Checkpointing & Keep last \\
GPU & A40 \\
Time & 3.5$\sim$6 hours \\
\bottomrule
\end{tabular}
\caption{Hyperparameters for training BERT on hidden units.}
\label{app:bert_hyper}
\end{table}

\subsection{WavEmbed}
For WavEmbed models, we apply the same hyperparameters in Table~\ref{app:wavembed_hyper}. 

\begin{table}
\centering
\begin{tabular}{lr}
\toprule
\textbf{Hyparameters} & \textbf{Value} \\ \midrule
Epoch & 1 \\
Total update steps & $\sim$13.5k \\
Optimizer & AdamW \\
Initial learning rate & 5e-4 \\
Effective batch size & 64 \\
FP 16 & False \\
Encoder initialization & HuBERT \\
Decoder initialization & Pretrained weights\\
Training task & Autoregressive LM \\
Loss & Cross-entropy loss \\
Checkpointing & Keep best \\
GPU & A40 \\
Time & $\sim$22 hours \\
\bottomrule
\end{tabular}
\caption{Hyperparameters for training WavEmbed models.}
\label{app:wavembed_hyper}
\end{table}

\subsection{Hu-TSDAE}
Table~\ref{app:hu-tsdae_hyper} presents the hyperparameters for all Hu-TSDAE models, which were trained for the following vocabulary sizes \{50, 100, 200, 1000, 3000, 5000, 8000, 12000\}. For vocabulary sizes larger than 200, they were learned from the hidden units with 100 clusters. 

\begin{table}
\centering
\begin{tabular}{lr}
\toprule
\textbf{Hyparameters} & \textbf{Value} \\ \midrule
Epoch & 3 \\
Total update steps & $\sim$10.4k \\
Optimizer & AdamW \\
Initial learning rate & 3e-5 \\
Effective batch size & 256 \\
FP 16 & False \\
Encoder initialization &  Pretrained weights\\
Decoder initialization & Same as encoder\\
Training task & Autoregressive LM \\
Loss & Cross-entropy loss \\
Checkpointing & Keep best \\
GPU & A40 \\
Time & 6$\sim$10.5 hours \\
\bottomrule
\end{tabular}
\caption{Hyperparameters for training Hu-TSDAE models.}
\label{app:hu-tsdae_hyper}
\end{table}

\subsection{Hu-SimCSE}
The general configurations were specified in Table~\ref{app:hu-simcse_hyper} and~\ref{tab:simcse-hu-grid}. For Hu-SimCSE models, the training was highly unstable. The loss dropped to a very low value in about 20 steps. Most models usually reached the best validation performance in less than 200 iterations, and then the performance began to drop severely. So early stoping was applied when the validation performance dropped for more than 80 steps. In our experiments, Hu-SimCSE models was highly sensitive to hyperparameters, so we tuned hyperparamers by carrying out grid search if effective batch size $\in$ \{64, 128, 256, 512\}, number of in-batch negative samples $\in$ \{63,127\} and learning rate $\in$ \{5e-5,1e-5\}. $\tau$ was set to 0.05 for all experiments.

\begin{table}
\centering
\begin{tabular}{lr}
\toprule
\textbf{Hyparameters} & \textbf{Value} \\ \midrule
Epoch & 1 \\
Optimizer & AdamW \\
FP 16 & False \\
Encoder initialization &  Pretrained weights\\
Training task & Contrastive learning \\
Loss & InfoNCE loss \\
Checkpointing & Keep best \\
Early stoping & true \\
GPU & A40 \\
Time & 5$\sim$28 minutes \\
\bottomrule
\end{tabular}
\caption{Hyperparameters for training Hu-SimCSE models.}
\label{app:hu-simcse_hyper}
\end{table}

\begin{table*}[]
\centering
\small
\begin{tabular}{lrrrrr}
\toprule
\textbf{Models} & \textbf{Clusters} & \textbf{Vocab.} & \textbf{Effective batch size} & \textbf{Learning rate} & \textbf{Num. Negatives} \\ \midrule
Hu-SimCSE & 50 & 50                    & 128 & 5e-5 & 63 \\
Hu-SimCSE & 100 & 100                   & 128 & 5e-5 & 63     \\
Hu-SimCSE & 200 & 200                   & 128 & 5e-5 & 63                                  \\\midrule
Hu-SimCSE & 100 & 1000       & 256 & 5e-5 & 127        \\
Hu-SimCSE & 100 & 3000      & 64 & 5e-5 & 63  \\
Hu-SimCSE & 100 & 5000      & 256 & 5e-5 & 127    \\
Hu-SimCSE & 100 & 8000      & 256 & 5e-5 & 63         \\
Hu-SimCSE & 100 & 12000     & 256 & 5e-5 & 127  \\
Hu-SimCSE & 100 & 20000 &  256 & 5e-5 & 127 \\ \midrule
Hu-SimCSE & 200 & 1000       & 128 & 5e-5 & 63 \\
Hu-SimCSE & 200 & 3000      &   512 & 5e-5 & 127       \\
Hu-SimCSE & 200 & 5000      &    128 & 5e-5 & 127      \\
Hu-SimCSE & 200 & 8000      &  256 & 5e-5 & 127    \\
Hu-SimCSE & 200 & 12000     &  256 & 5e-5 & 127 \\
Hu-SimCSE & 200 & 20000 & 128 & 5e-5 & 63 \\ \bottomrule
\end{tabular}
\caption{Final hyperparameters for Hu-SimCSE models.}
\label{tab:simcse-hu-grid}
\end{table*}

\subsection{S-HuBERT}
The hyperparameters for S-HuBERT were shown in Table~\ref{app:distil_hyper}. The text encoder was pretrained text embeddings including Hu-SimCSE, Hu-TSDAE, SimCSE, and TSDAE. The weights of text encoder were frozen throughout training. 

\begin{table}
\centering
\begin{tabular}{lr}
\toprule
\textbf{Hyparameters} & \textbf{Value} \\ \midrule
Epoch & 1 \\
Total update steps & $\sim$4k \\
Optimizer & AdamW \\
Initial learning rate & 1e-4 \\
Effective batch size & 192\\
Temperature & 0.05 \\
Memory bank & 256 \\
FP 16 & False \\
Speech Encoder &  HuBERT \\
Text Encoder & Pretrained weights\\
Training task & Contrastive learning \\
Loss & InfoNCE \\
Checkpointing & Keep best \\
GPU & A40 \\
Time & $\sim$21 hours \\
\bottomrule
\end{tabular}
\caption{Hyperparameters for S-HuBERT.}
\label{app:distil_hyper}
\end{table}

\subsection{Unsupervised text embeddings: SimCSE and TSDAE}

For both SimCSE and TSDAE, we trained both text-based models using \texttt{sentence\_transformer} library \cite{reimers-gurevych-2019-sentence}, as it provides a easy-to-use interface. They were trained on the full texts of Common Voice English training subset. Models were initialized with either \texttt{bert-base-uncased} or \texttt{roberta-base-uncased}.
The following hyperparameters were used during model pretraining. 
\begin{itemize}
    \item Epoch: 2 (for SimCSE) or 3 (for TSDAE)
    \item Initial learning rate: 1e-4
    \item Effective batch size: 64 
\end{itemize}
Otherwise we used the default settings in \texttt{sentence\_transformer} for hyperparameters not mentioned above. 

In our experiments, it shows that using \texttt{bert-base-uncased} as initial model perform better (see Table~\ref{tab:wavembed}, Table~\ref{tab:distil}, and Table~\ref{tab:extended}).

\begin{table*}[ht]
\centering
\small
\begin{tabular}{lrrrrrrrr}
\toprule
\multicolumn{1}{c}{\multirow{2}{*}{\bf Models}} & \multicolumn{5}{c}{\bf Synthetic speech}  & \multicolumn{3}{c}{\bf Natural Speech} \\  \cmidrule(lr){2-6} \cmidrule(lr){7-9} 
 & \textbf{STS12} & \textbf{STS13} & \textbf{STS14} & \textbf{STS15} & \textbf{STS16} & \textbf{STS}  & \textbf{CVS-dev}  & \textbf{CVS-test}  \\ \toprule
\multicolumn{8}{l}{\it Unsupervised text embeddings} \\ \midrule
SimCSE-unsup-RoBERTa (CV text) & 52.7 & 70.0 & 63.2 & 72.2 & 72.8 &66.6 & 86.5  & 86.9\\
SimCSE-unsup-RoBERTa (ASR text) & 53.4 & 67.0  & 62.2  & 72.7 & 71.8 & 67.5 & 86.9  & 87.1\\
TSDAE-RoBERTa (CV text) & 45.0 & 59.2 & 54.7  & 65.3 & 70.9 & 57.1 & 88.0  & 86.7\\
TSDAE-RoBERTa (ASR text) & 44.9 & 61.1  & 55.7  & 68.2 & 72.8 & 60.5 & 88.2  & 86.8\\
\bottomrule
\end{tabular}
\caption{Evaluation results based on the Spearman's rank correlation. All models (e.g., HuBERT, RoBERTa, BERT) refer to the 12-layer base models.}
\label{tab:extended}
\end{table*}


\section{Ablation analysis}\label{app:ablation}

\subsection{Deletion ratio of Hu-TSDAE}
\label{app:deletion_rate}
In the original implementation of TSDAE, the input data were subject to random deletion with a ratio of 0.6, as this augmentation led to the best evaluation performance \cite{wang-etal-2021-tsdae-using}. However, we found that training TSDAE on hidden units did not benefit from random deletion. In Table~\ref{fig:del}, the Spearman's rank correlation was the highest when no deletion was applied.

\begin{table}[tbh]
\centering
\begin{tabular}{cr}
\toprule
\textbf{Deletion ratio} & \multicolumn{1}{c}{\textbf{CVS-dev}} \\ \midrule
0                       & \textbf{52.7}                                 \\
0.1                     & 49.2                                 \\
0.2                     & 42.9                                 \\
0.3                     & 40.5                                 \\
0.4                     & 29.4                                 \\
0.5                     & 23.7                                 \\
0.6                     & 18.2                                 \\ \bottomrule
\end{tabular}
\caption{The impact of deletion rate on the Spearman's rank correlations ($\times100$) of CVS-dev evaluation. Here the base model was the Hu-TSDAE model, which was trained on pseudo-texts with 100 clusters, which were further tokenized to 8000 vocabulary tokens.}
\label{fig:del}
\end{table}

\subsection{Impact of hidden units}\label{app:hu_impact}

The impact of vocabulary sizes on the validation performance was illustrated in Table~\ref{tab:tsdae-hu-impact} and~\ref{tab:simcse-hu-impact}. The vocabularies were induced from the whole Common Voice English corpus of hidden units. All models were tested on the CVS development set, which was also discretized into hidden units. 

Both tables suggest that re-tokenizing hidden units into a higher level of sub-word units generally improve performance for directly modeling hidden units without speech involved. Generally speaking, smaller vocabulary size usually leads to higher performance for Hu-TSDAE, as 1000 is the best vocabulary size. However, for Hu-SimCSE, having a vocabulary size that is not too large nor too small is more beneficial for training (8000 was the optimal size in our experiments).

\begin{table}
\small
\begin{tabular}{lrrr}
\toprule
\multicolumn{1}{c}{\textbf{Models}} & \multicolumn{1}{c}{\textbf{Clusters}} & \multicolumn{1}{c}{\textbf{Vocab. Size}} & \multicolumn{1}{c}{\textbf{CVS-dev}} \\ \midrule
Hu-TSDAE & 50 & 50                    & 49.9                                 \\
Hu-TSDAE & 100 & 100                   & 55.8                                 \\
Hu-TSDAE & 200 & 200                   & 56.9                                 \\\midrule
Hu-TSDAE & 100 & 1000       & \textbf{58.4}                                 \\
Hu-TSDAE & 100 & 3000      & 57.5                                 \\
Hu-TSDAE & 100 & 5000      & 56.9                                 \\
Hu-TSDAE & 100 & 8000      & 54.3                                 \\
Hu-TSDAE & 100 & 12000     & 52.8                                 \\ \bottomrule
\end{tabular}
\caption{For Hu-TSDAE models, the impact of the number of hidden-unit clusters and the vocabulary size on the Spearman's correlation ($\times100$). Results were based on the hidden-unit transcriptions of CVS development set.}
\label{tab:tsdae-hu-impact}
\end{table}

\begin{table}
\centering
\small
\begin{tabular}{lrrr}
\toprule
\multicolumn{1}{c}{\textbf{Models}} & \multicolumn{1}{c}{\textbf{Clusters}} & \multicolumn{1}{c}{\textbf{Vocab. Size}} & \multicolumn{1}{c}{\textbf{CVS-dev}} \\ \midrule
Hu-SimCSE & 50 & 50                    & 30.7                                 \\
Hu-SimCSE & 100 & 100                   & 31.9                                 \\
Hu-SimCSE & 200 & 200                   & 31.9                                 \\\midrule
Hu-SimCSE & 100 & 1000       & 35.2                                 \\
Hu-SimCSE & 100 & 3000      & 37.1                                 \\
Hu-SimCSE & 100 & 5000      & 37.3                                 \\
Hu-SimCSE & 100 & 8000      & \textbf{42.5}                                 \\
Hu-SimCSE & 100 & 12000     & 35.6  \\
Hu-SimCSE & 100 & 20000 & 36.8 \\ \midrule
Hu-SimCSE & 200 & 1000       & 36.8 \\
Hu-SimCSE & 200 & 3000      &   34.5       \\
Hu-SimCSE & 200 & 5000      &   36.1      \\
Hu-SimCSE & 200 & 8000      & \textbf{39.4}     \\
Hu-SimCSE & 200 & 12000     &  38.5 \\
Hu-SimCSE & 200 & 20000 & 33.7 \\ \bottomrule
\end{tabular}
\caption{For Hu-SimCSE models, the impact of the number of hidden units and the vocabulary size on the evaluation Spearman's rank correlation ($\times100$). Results were based on the hidden-unit transcriptions of CVS development set.}
\label{tab:simcse-hu-impact}
\end{table}

\begin{table*}
\centering
\small
\begin{tabular}{lrrrrr}
\toprule
\textbf{Models} & \textbf{Pooling} & \textbf{Loss} & \textbf{STS} & \textbf{CVS-dev} & \textbf{CVS-test} \\ \midrule

S-HuBERT (\textit{\textcolor{NavyBlue}{SimCSE-unsup-RoBERTa}}) & CLS & MSE & 68.6 & 79.5 & 77.9 \\
S-HuBERT (\textit{\textcolor{NavyBlue}{SimCSE-unsup-RoBERTa}}) & Self-attention & MSE & 68.6 & \textbf{80.7} & 80.3 \\
S-HuBERT (\textit{\textcolor{NavyBlue}{SimCSE-unsup-RoBERTa}}) & CLS & InfoNCE & 67.8 & 80.0 & 79.1 \\
S-HuBERT (\textit{\textcolor{NavyBlue}{SimCSE-unsup-RoBERTa}}) & Self-attention & InfoNCE & \textbf{70.2} & 79.8 & \textbf{81.6} \\\bottomrule
\end{tabular}
\caption{Impact of different combination of pooling methods and loss functions on evaluation performance. The italicized blue font suggests that results or checkpoints were directly taken from \citet{gao-etal-2021-simcse}. Models in parentheses are teacher models whereas models outside parentheses are student models.}
\label{tab:pooling-and-loss}
\end{table*}

\begin{table*}[]
\centering
\small
\begin{tabular}{lrrrrrr}
\toprule
\textbf{Models} & \textbf{Encoder Init.} & \textbf{Decoder Init.} &  \textbf{STS} & \textbf{CVS-dev} & \textbf{CVS-test} \\ \midrule
WavEmbed & HuBERT-base-ls960 &  GPT-2-base &  52.4       &    76.5         &   77.4           \\ 
WavEmbed & HuBERT-base-ls960 &BERT-base-uncased    &  55.4     &   \textbf{78.4}          &     78.7         \\ 
WavEmbed & HuBERT-base-ls960 &RoBERTa-base  &    \textbf{58.6}    &      78.1       &  \textbf{79.4}            \\ 
WavEmbed & HuBERT-base-ls960 & Random  &    37.5 & 74.6 & 76           \\
WavEmbed & Random & Random  &   28.9 & 60.4 &53.4            \\\midrule
WavEmbed - 100 C   & HuBERT-base-ls960 &  BERT pretrained  on hidden units     & 50.0 & 65.6 & 61.2 \\ 
WavEmbed - 100 C & HuBERT-base-ls960 & Random     & 46.8 &
67.5 & 61.9 \\ 
\bottomrule
\end{tabular}
\caption{Impact of weight initialization on the performance of WavEmbed. These results show that initializing with pretrained weights is important for improving performance.}
\label{tab:weights}
\end{table*}

\subsection{Pooling methods and loss functions}\label{app:pooling-and-loss}
Here we compared how different pooling methods and loss functions can impact model performance. For loss functions, we compared InfoNCE and mean square errors (MSE), both of which are commonly used for knowledge distillation. 

For pooling methods, \citet{khurana2022samu} has shown that self-attention pooling is better than mean or max pooling, so we did not repeat the same experiments here. 

Two pooling methods for HuBERT models were compared: self-attention pooling and CLS pooling. The CLS pooling was proposed by \citet{duquenne2021multimodal}, in which a CLS token (a vector of ones) is inserted before speech feature extractors before feeding them into the transformer part of HuBERT. Then the last hidden state of the CLS token, or the first embedding of the last hidden states, was used as the representation of the whole speech signal.
This is similar to pretrained text transformers like BERT \cite{devlin-etal-2019-bert} and RoBERTa \cite{liu2019roberta}. Here we inserted a learnable embedding for the CLS token at the beginning of the sequence. Results in Table~\ref{tab:pooling-and-loss} suggest that self-attention pooling and the InfoNCE loss work best.

\subsection{Weight initialization}\label{app:weight}
Here we compared how initalizing with pretrained weights affect model performance. For WavEmbed, we initialized the WavEmbed decoder with different model weights, including pretrained weights or random weights. When initializing a decoder with encocders such as BERT and RoBERTa, all cross-attention layers were reinitialized. Despite this, initializing the decoder with pretrained weights can help improve the performance, as shown in Table~\ref{tab:weights}. 


\section{Visualizing predictions}\label{app:vis-pred}
\begin{figure*}
    \centering
    \includegraphics[width=0.48\textwidth]{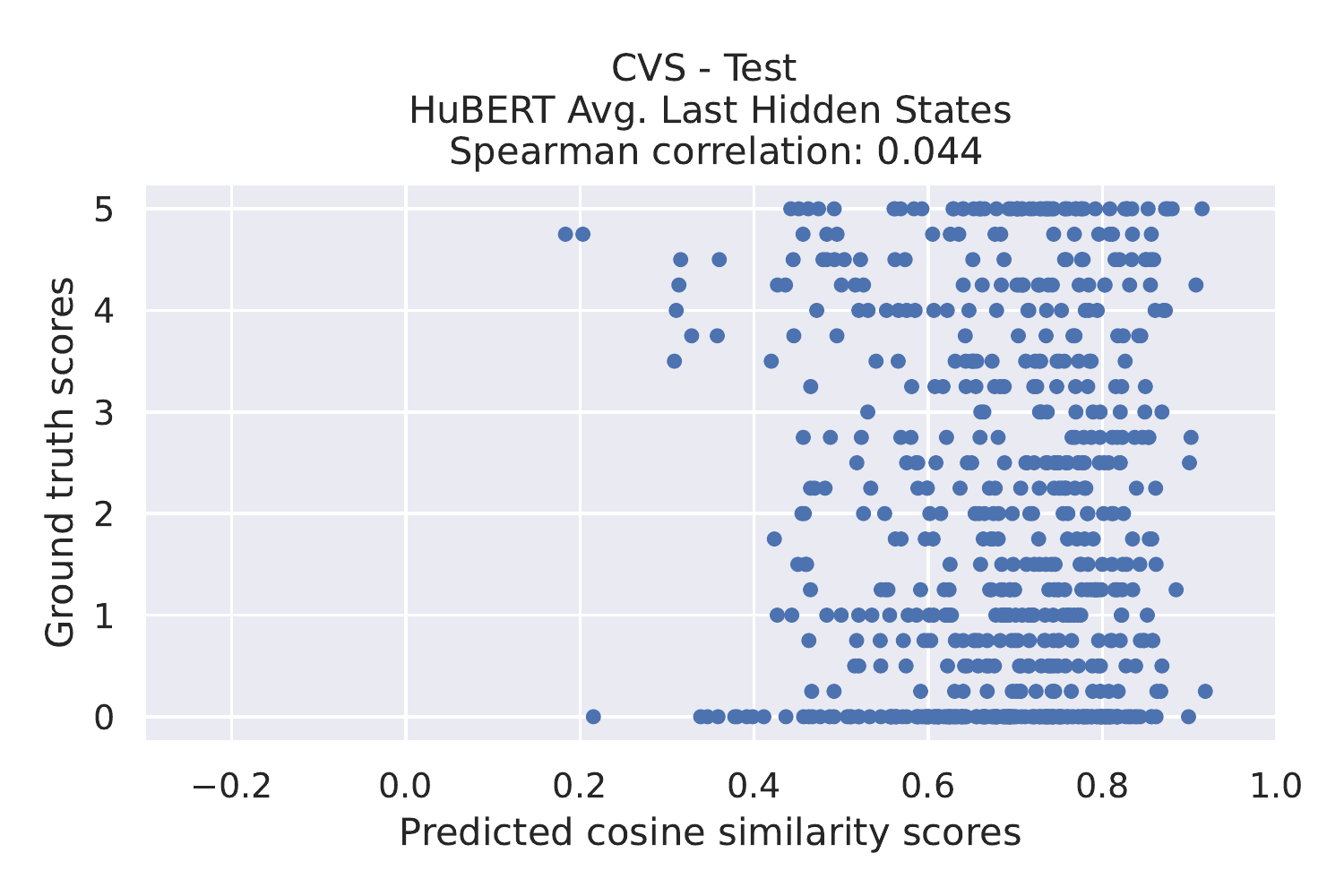}
    \includegraphics[width=0.48\textwidth]{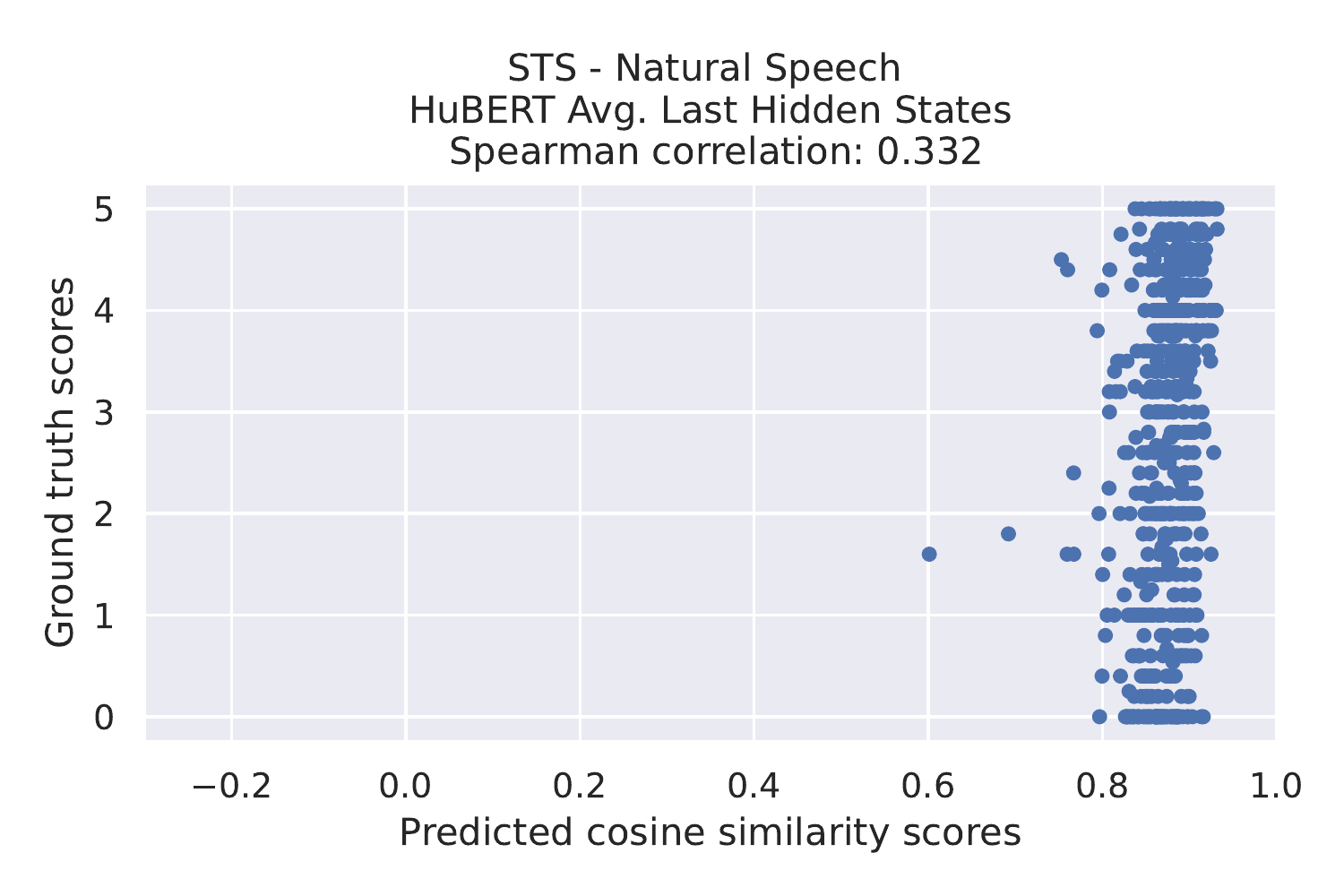}
    \includegraphics[width=0.48\textwidth]{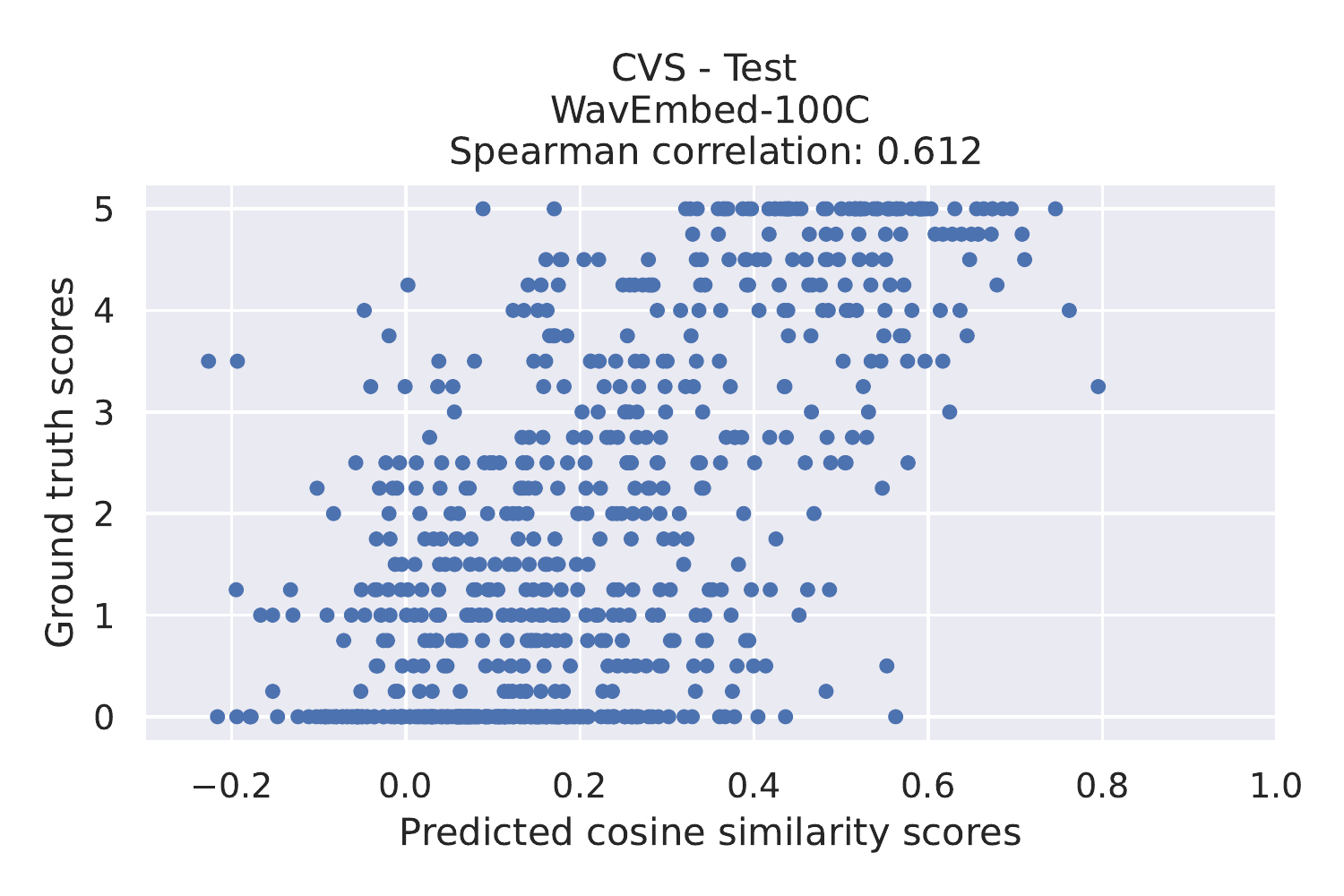}
    \includegraphics[width=0.48\textwidth]{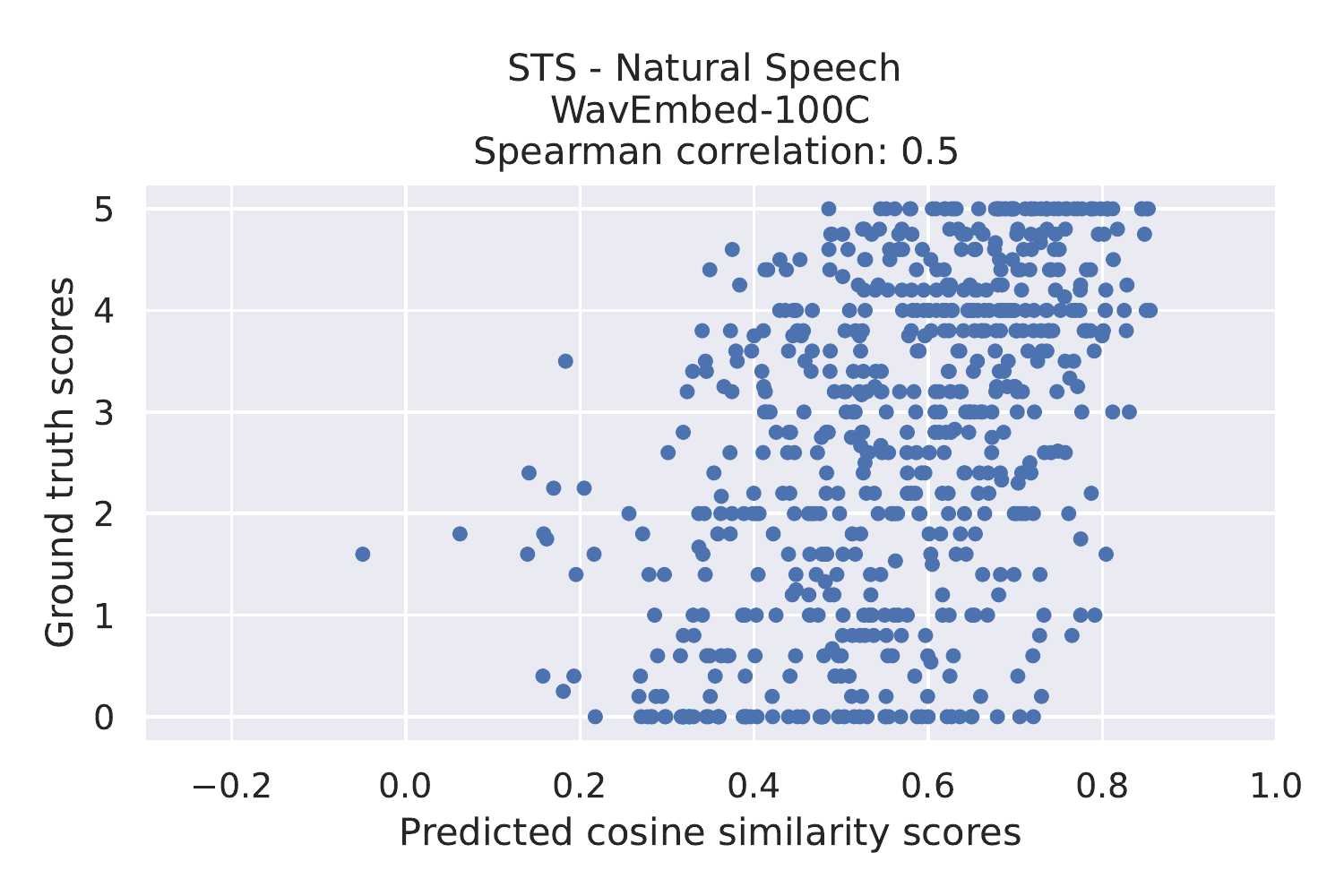}
    \includegraphics[width=0.48\textwidth]{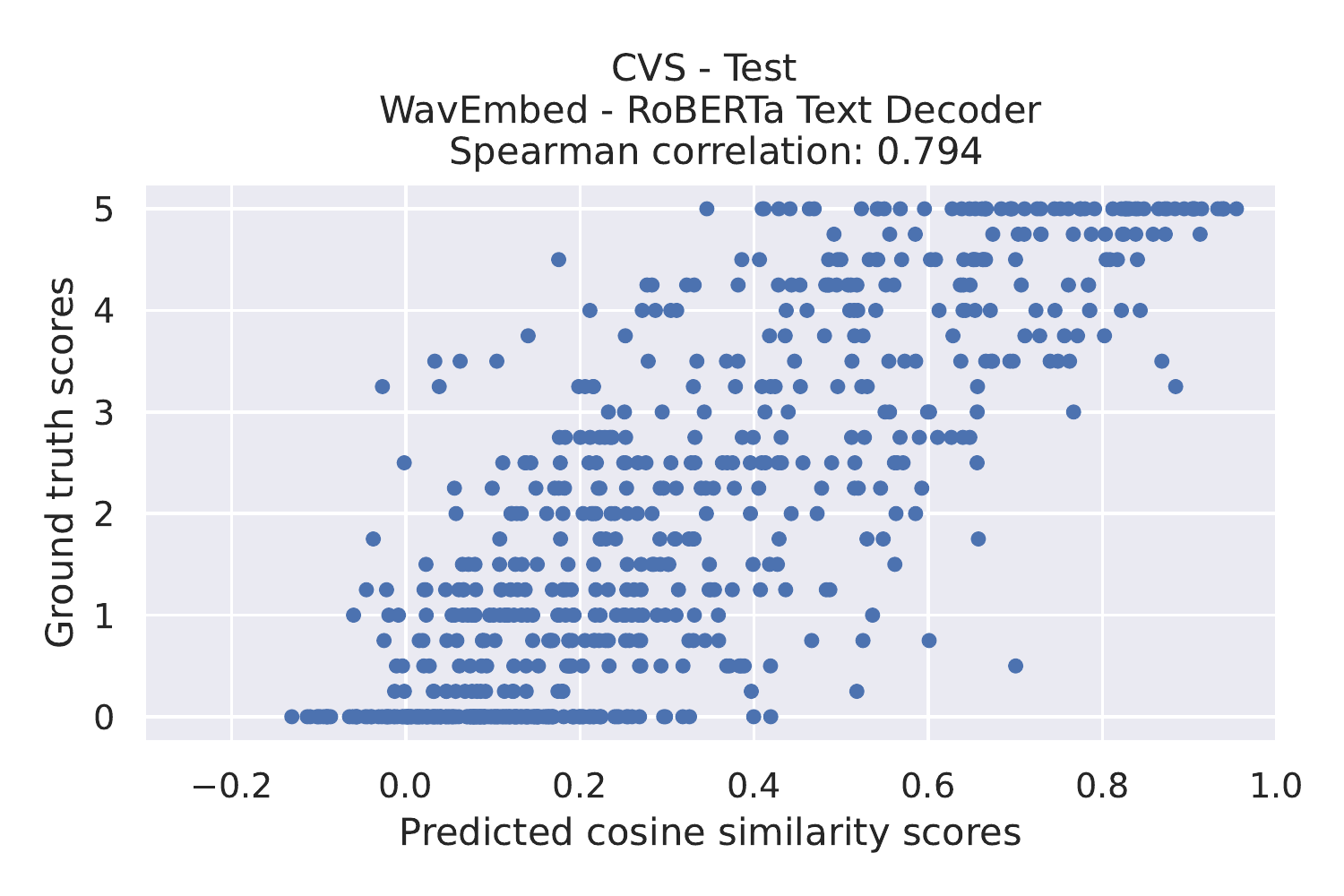}
    \includegraphics[width=0.48\textwidth]{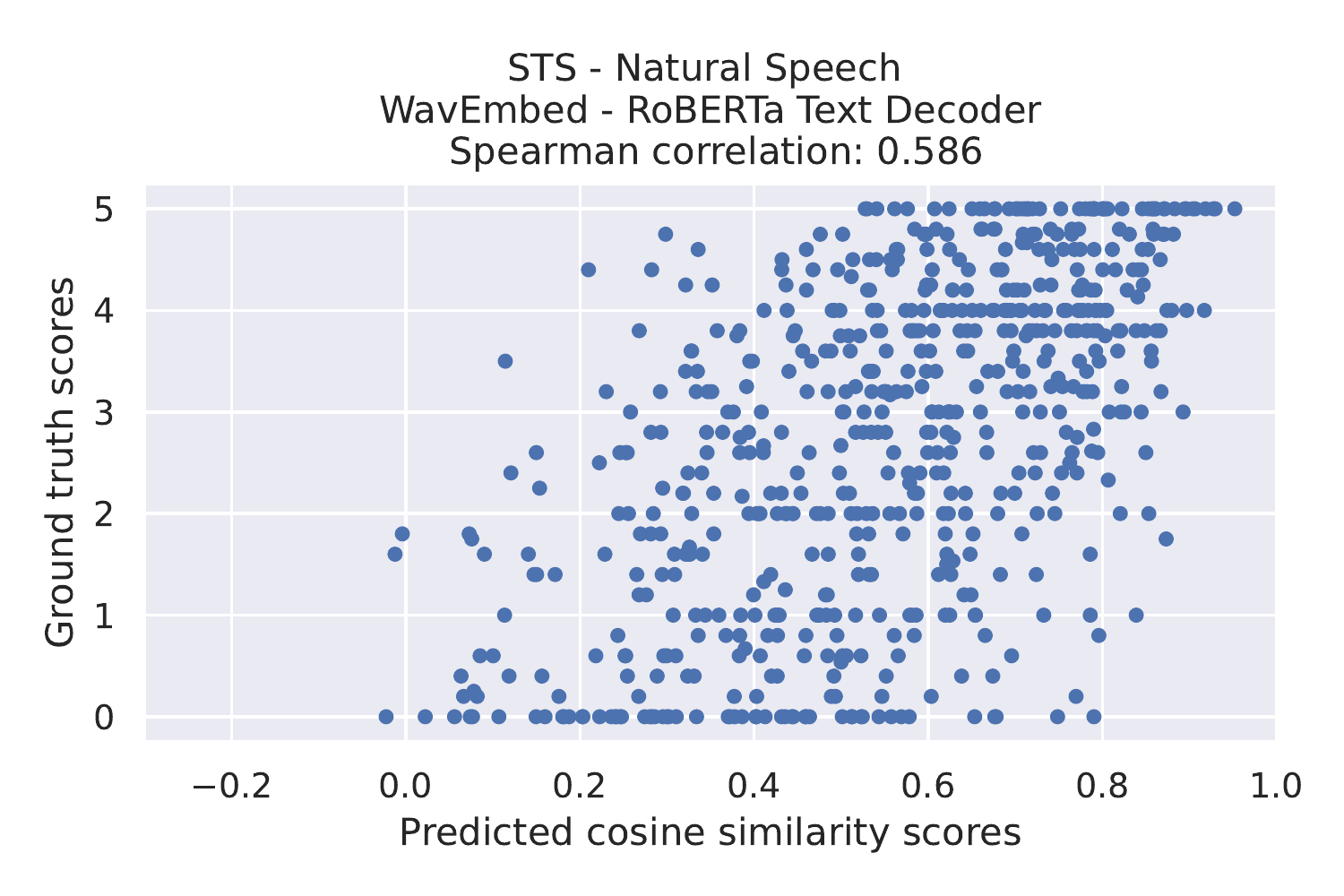}
    \caption{Plots of predicted semantic similarity scores by different WavEmbed models against ground truth scores.}
    \label{fig:wavembed-corr}
\end{figure*}

\begin{figure*}
    \centering
    \includegraphics[width=0.48\textwidth]{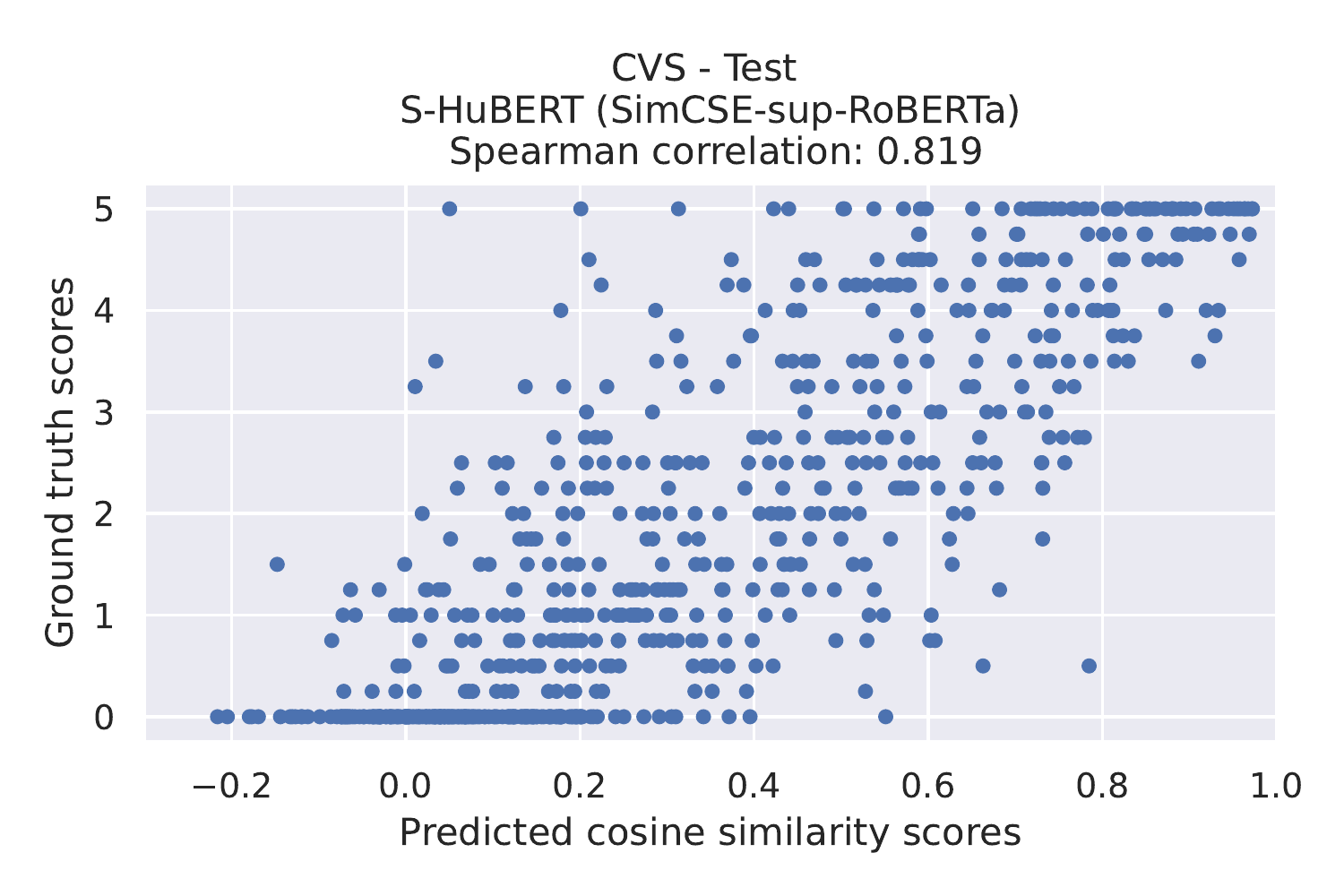}
    \includegraphics[width=0.48\textwidth]{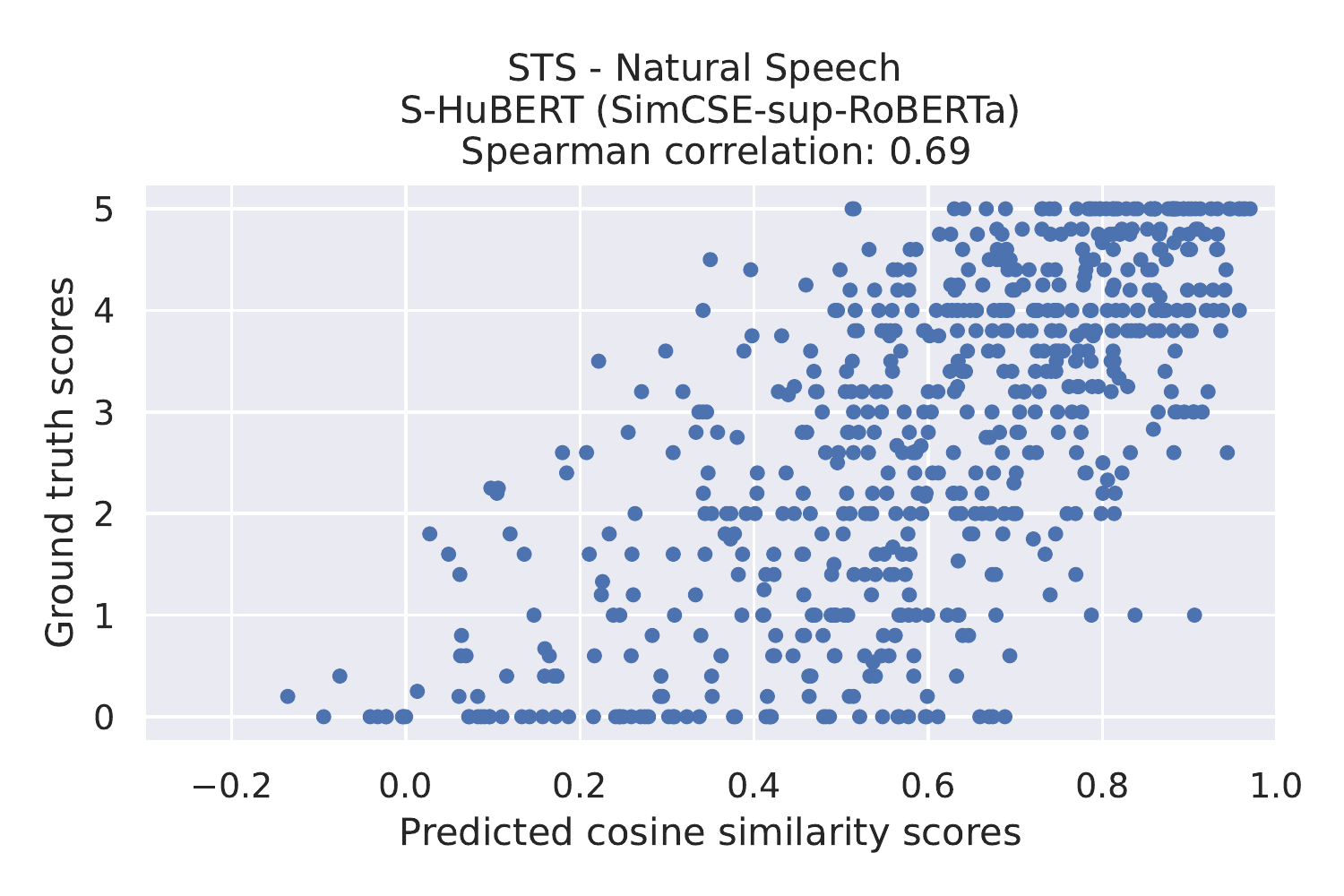}
    \includegraphics[width=0.48\textwidth]{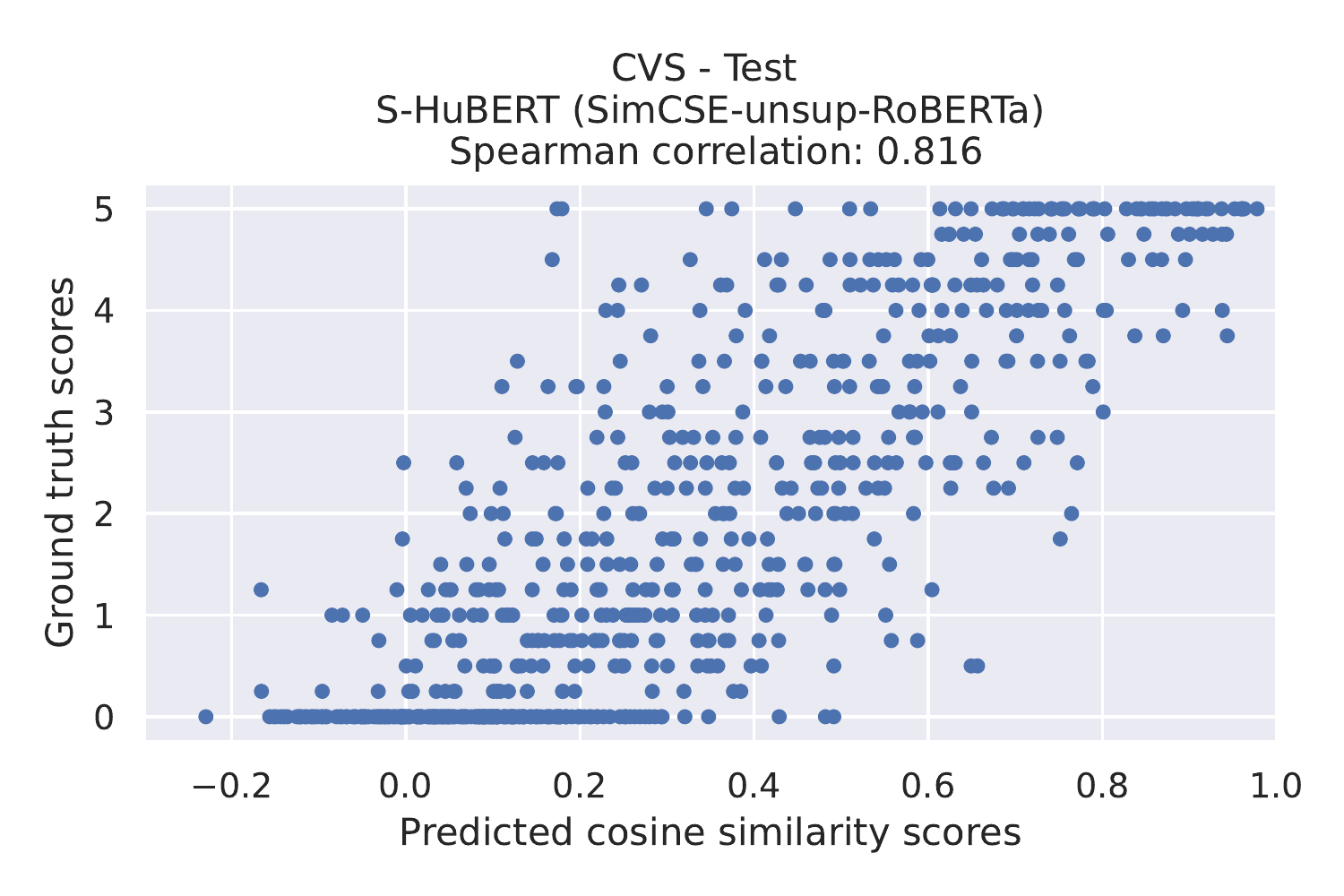}
    \includegraphics[width=0.48\textwidth]{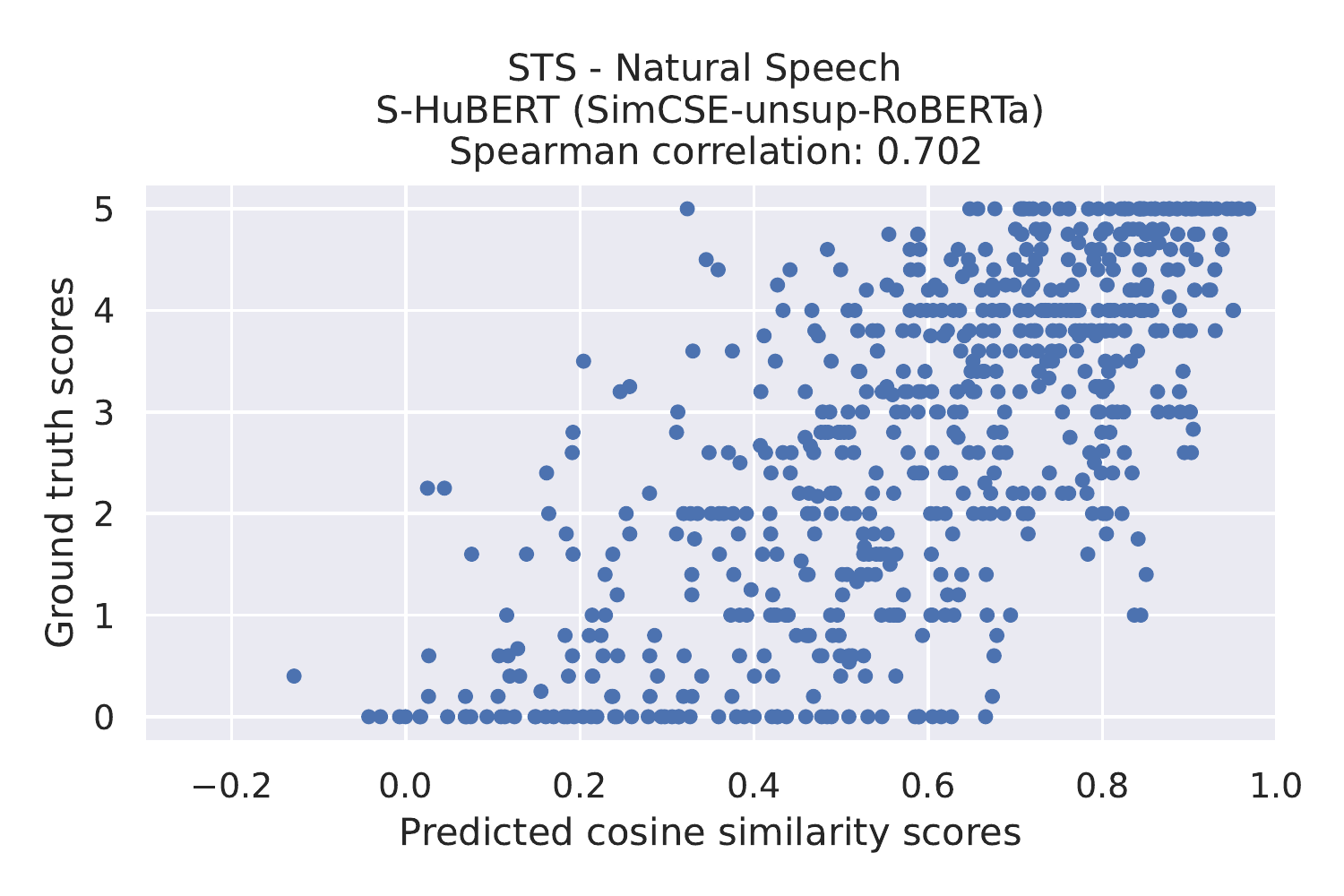}
    \includegraphics[width=0.48\textwidth]{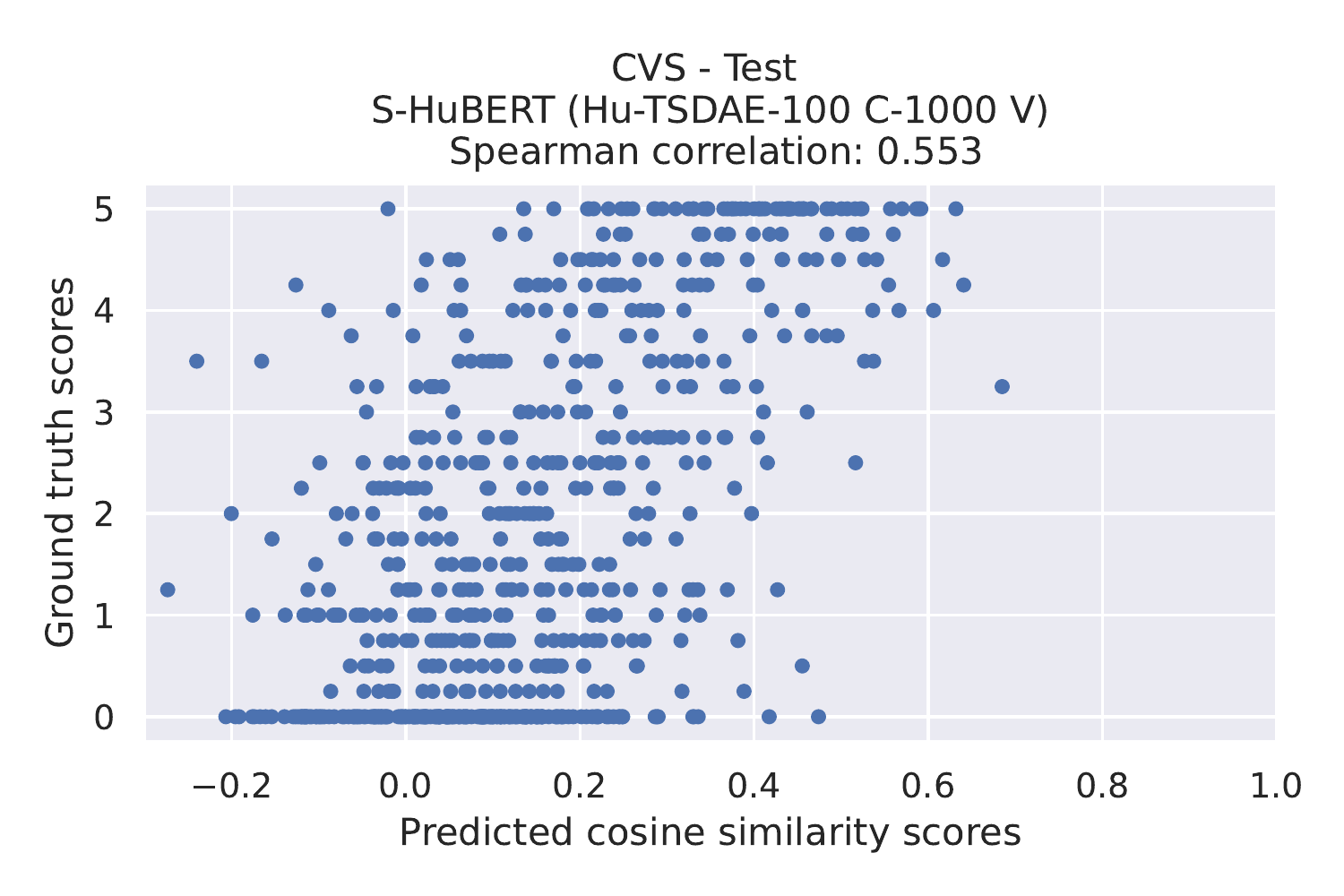}
    \includegraphics[width=0.48\textwidth]{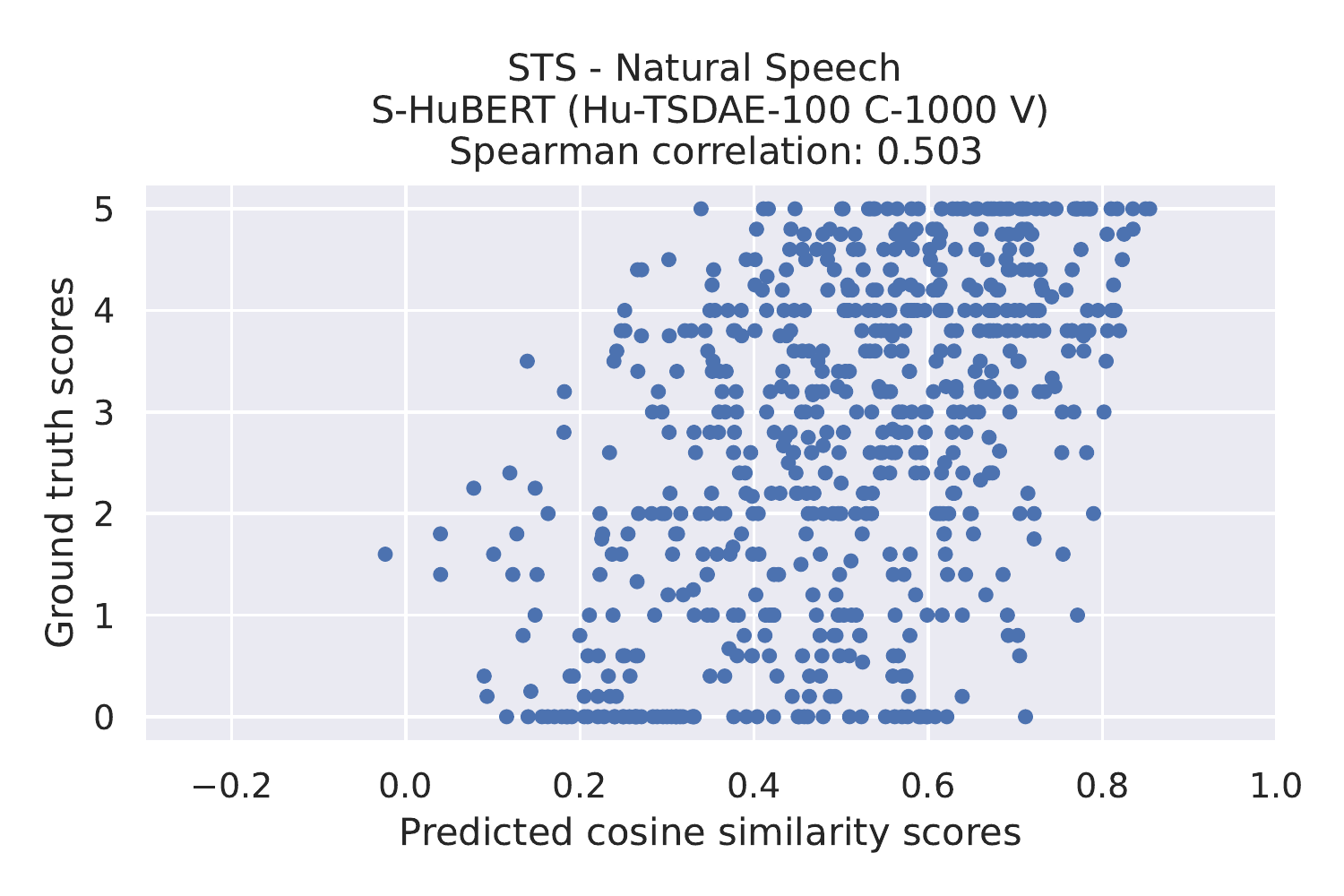}
    \includegraphics[width=0.48\textwidth]{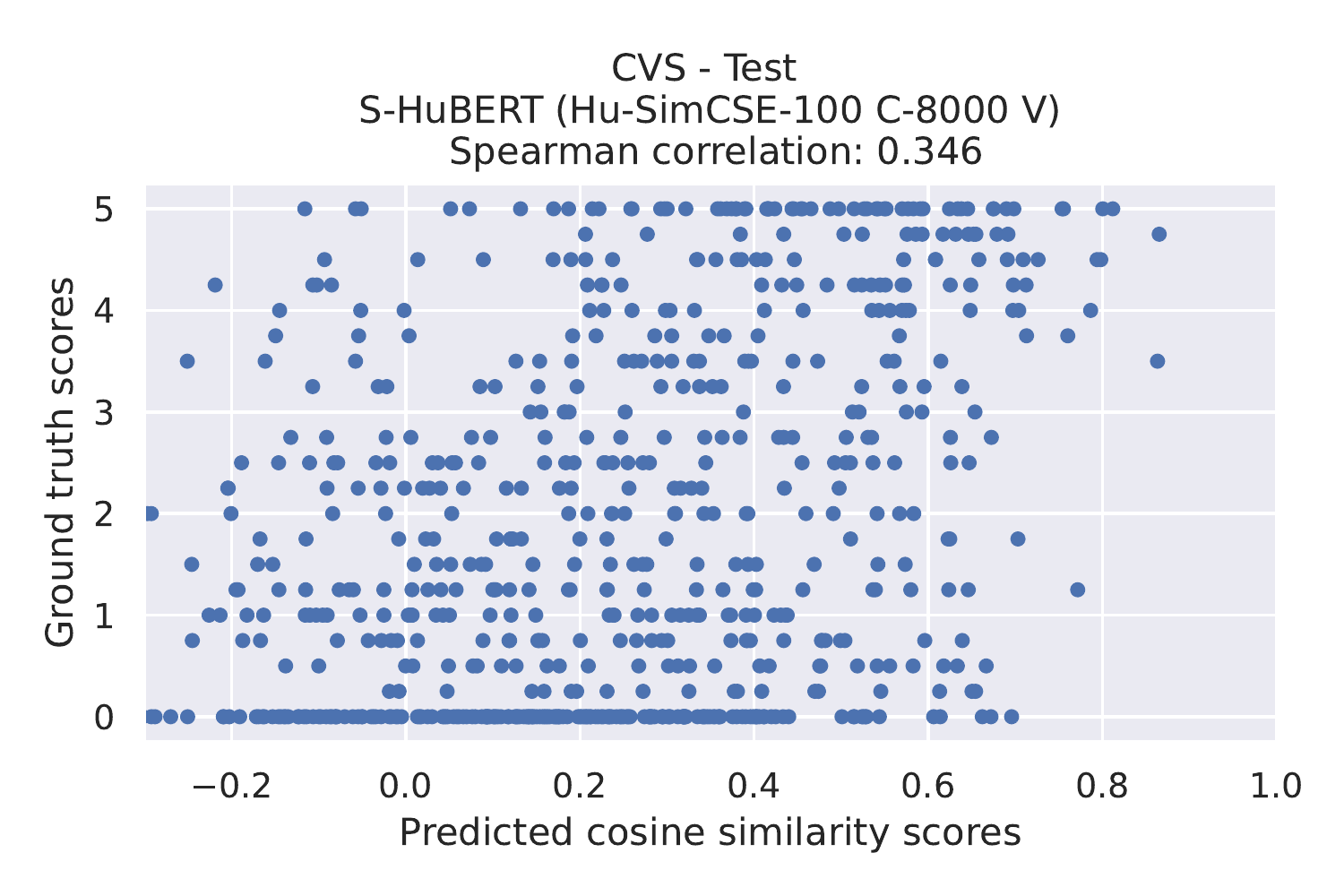}
    \includegraphics[width=0.48\textwidth]{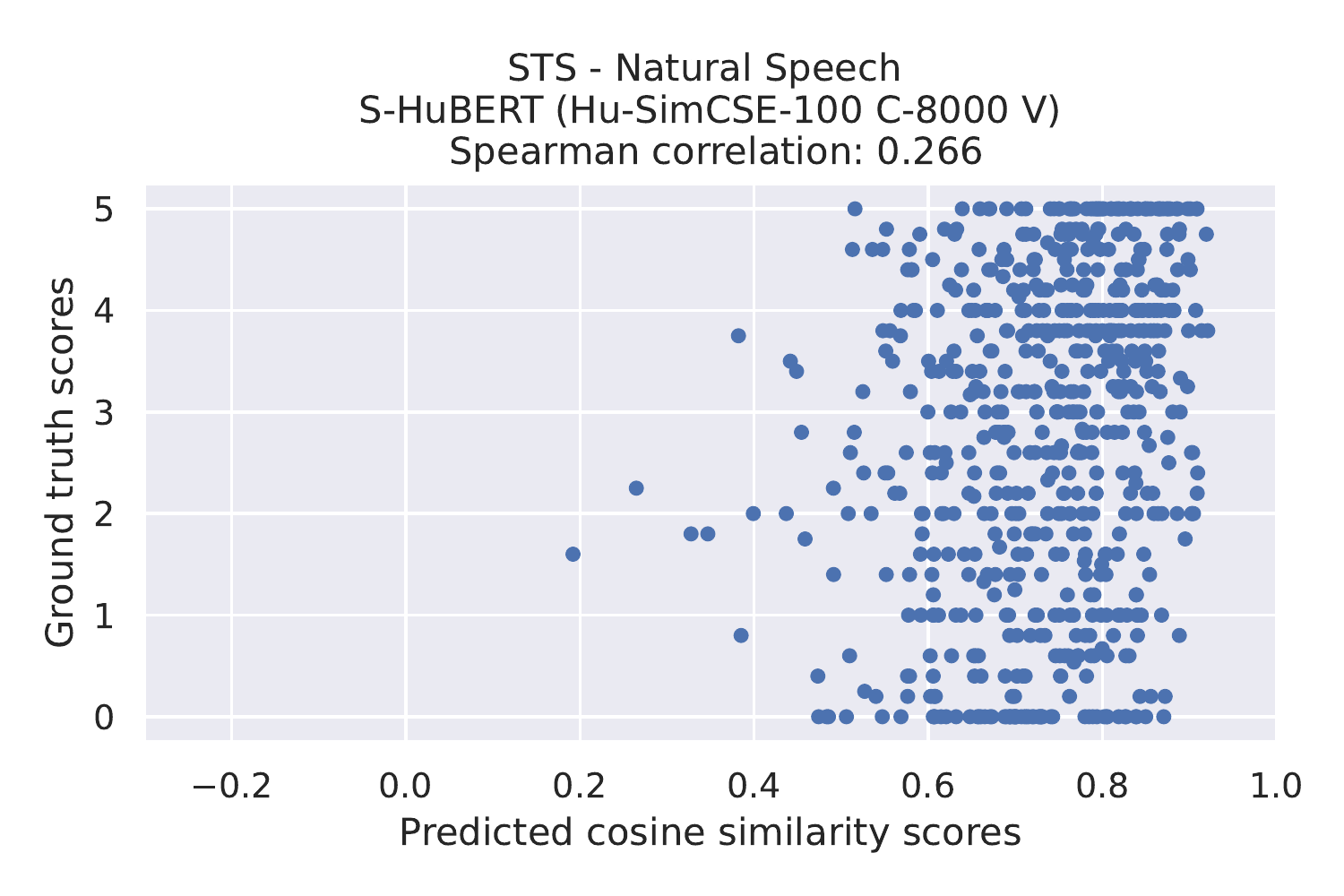}
    \caption{Plots of predicted semantic similarity scores by different S-HuBERT models against ground truth scores. }
    \label{fig:shubert-corr}
\end{figure*}

Details predictions of selected models were shown at Figure~\ref{fig:wavembed-corr} and~\ref{fig:shubert-corr}.

\end{document}